\documentclass{article}

    \usepackage[final, nonatbib]{neurips_2022}

\usepackage[utf8]{inputenc} 
\usepackage[T1]{fontenc}    
\usepackage{hyperref}       
\usepackage{url}            
\usepackage{booktabs}       
\usepackage{amsfonts}       
\usepackage{nicefrac}       
\usepackage{microtype}      

\usepackage{graphicx}
\usepackage{tikz}
\usepackage{comment}
\usepackage{amsmath,amssymb} 
\usepackage{xcolor}

\usepackage{graphicx}
\usepackage{booktabs}
\usepackage{latexsym}

\usepackage{url}
\usepackage{booktabs}
\usepackage{multirow}
\usepackage{array, caption, floatrow, makecell, booktabs}
\usepackage{wrapfig}
\usepackage{floatrow}
\newfloatcommand{capbtabbox}{table}[][\FBwidth]

\usepackage{enumitem}
\usepackage{accents}
\usepackage{xspace,mfirstuc,tabulary}
\usepackage{tabu}
\usepackage{wrapfig}
\usepackage{graphicx, amsmath, amssymb, caption, subcaption, multirow, overpic, textpos}
\usepackage{todonotes}

\newcommand\blfootnote[1]{%
  \begingroup
  \renewcommand\thefootnote{}\footnote{#1}%
  \addtocounter{footnote}{-1}%
  \endgroup
}
\definecolor{airforceblue}{rgb}{0.36, 0.54, 0.66}
\definecolor{amaranth}{rgb}{0.9, 0.17, 0.31}
\makeatletter
\newcommand*\iftodonotes{\if@todonotes@disabled\expandafter\@secondoftwo\else\expandafter\@firstoftwo\fi}  
\makeatother

\usepackage{pifont}
\newcommand{\cmark}{\ding{51}}%
\newcommand{\xmark}{\ding{55}}%

\usepackage{amsmath}
\usepackage{amsfonts}
\usepackage{amssymb}

\usepackage{wrapfig}

\newcommand{\RN}[1]{%
	\textup{\lowercase\expandafter{\it \romannumeral#1}}%
}

\usepackage{xcolor}
\definecolor{mygreen}{HTML}{3cb44b}
\definecolor{skyblue}{HTML}{beffff}
\definecolor{lightgreen}{HTML}{90ee90}

\newcommand{\beq}{\vspace{0mm}\begin{equation}}
\newcommand{\eeq}{\vspace{0mm}\end{equation}}
\newcommand{\beqs}{\vspace{0mm}\begin{eqnarray}}
\newcommand{\eeqs}{\vspace{0mm}\end{eqnarray}}
\newcommand{\barr}{\begin{array}}
\newcommand{\earr}{\end{array}}

\usepackage{caption}

\newcommand{\Lcal}{\mathcal{L}}

\definecolor{Gray}{gray}{0.93}
\definecolor{Graylight}{gray}{0.95}
\definecolor{Grayheavy}{gray}{0.90}

\usepackage{colortbl}
\definecolor{Gray}{gray}{0.93}

\usepackage{tabu}
\usepackage{array, caption, floatrow, makecell, booktabs}
\usepackage{wrapfig}
\usepackage{floatrow}
\usepackage{comment}
\usepackage{float}
\usepackage{wrapfig,lipsum,booktabs}
\newcommand{\our}{GLIPv2\xspace}
\newcommand{\ourT}{GLIPv2-T\xspace}
\newcommand{\ourB}{GLIPv2-B\xspace}
\newcommand{\ourH}{GLIPv2-H\xspace}
\newcommand{\std}[1]{\tiny{$\pm$#1}}

\title{\our: Unifying Localization and VL Understanding}

\author{
\textbf{\normalsize{Haotian Zhang$^{*1\dagger}$, Pengchuan Zhang$^{*2\dagger\spadesuit}$, Xiaowei Hu$^{3}$,
Yen-Chun Chen$^{3}$, Liunian Harold Li$^{4\dagger}$}} \\ 
\textbf{\normalsize{Xiyang Dai$^{3}$, Lijuan Wang$^{3}$, Lu Yuan$^{3}$, Jenq-Neng Hwang$^{1}$, Jianfeng Gao$^{3}$}} \\
\normalsize{$^1$University of Washington, $^2$Meta AI, $^3$Microsoft, $^{4}$UCLA}\\
\texttt{\scriptsize{ \{haotiz,hwang\}@uw.edu,pengchuanzhang@fb.com,liunian.harold.li@cs.ucla.edu,}} \\
\texttt{\scriptsize{ \{Xiaowei.Hu,Yen-Chun.Chen,Xiyang.Dai,lijuanw,luyuan,jfgao\}@microsoft.com}}
}

\begin{document}

\maketitle

\begin{abstract}
    We present \our, a grounded VL understanding model, that serves both localization tasks (e.g., object detection, instance segmentation) and Vision-Language (VL) understanding tasks (e.g., VQA, image captioning). \our elegantly unifies localization pre-training and Vision-Language Pre-training (VLP) with three pre-training tasks: phrase grounding as a VL reformulation of the detection task, region-word contrastive learning as a novel region-word level contrastive learning task, and the masked language modeling. This unification not only simplifies the previous multi-stage VLP procedure but also achieves mutual benefits between localization and understanding tasks. Experimental results show that a single \our model (all model weights are shared) achieves near SoTA performance on various localization and understanding tasks. The model also shows (1) strong zero-shot and few-shot adaption performance on open-vocabulary object detection tasks and (2) superior grounding capability on VL understanding tasks. Code is released at \url{https://github.com/microsoft/GLIP}. \blfootnote{$^*$The two authors contributed equally. $^\dagger$Work done at Microsoft Research. $^\spadesuit$ Corresponding author.}

\end{abstract}

\vspace{-6mm}
\section{Introduction}
\label{sec:intro}

Recently, a general interest arises in building general-purpose vision systems \cite{gupta2021towards,hu2021unit,yang2021crossing,lu2022unified}, also called vision foundation models \cite{bommasani2021opportunities,yuan2021florence}, that solve various vision tasks simultaneously, such as image classification~\cite{krizhevsky2012imagenet}, object detection~\cite{lin2014microsoft}, and Visual-Language (VL) understanding~\cite{VQA,chen2015microsoft,kiros2014unifying}. 
Of particular interest, is the unification between \emph{localization} tasks (e.g., object detection~\cite{lin2014microsoft} and segmentation~\cite{caesar2018coco,gupta2019lvis}) and VL \emph{understanding} tasks (e.g., VQA~\cite{VQA} and image captioning~\cite{chen2015microsoft}). 
Localization pre-training benefits VL tasks~\cite{Anderson2017up-down,zhang2021vinvl}, and the ``localization->VLP'' two-stage pre-training procedure~\cite{lu2019vilbert,tan2019lxmert,chen2019uniter,su2019vl,li2019visualbert,li2019unicoder,zhou2019unified,li2020oscar,li2020unsupervised} is the common practice in VL community. A long-standing challenge is the unification of localization and understanding, which aims at \emph{mutual} benefit between these two kinds of tasks, simplified pre-training procedure, and reduced pre-training cost. 

However, these two kinds of tasks appear to be dramatically different: localization tasks are vision-only and require fine-grained output (e.g., bounding boxes or pixel masks), while VL understanding tasks emphasize fusion between two modalities and require high-level semantic outputs (e.g., answers or captions).

\cite{gupta2021towards,hu2021unit,yang2021crossing} have made early attempts at unifying these tasks in a straightforward multi-task manner, where a low-level visual encoder is shared across tasks, and two separate high-level branches are designed for localization and VL understanding, respectively. The localization tasks are still vision-only and do not benefit from the rich semantics in vision-language data. As a result, such unified models see the marginal mutual benefit or even performance degradation~\cite{hu2021unit} compared with task-specific models. 

\begin{figure}[t]
    \centering
    \includegraphics[width=\linewidth]{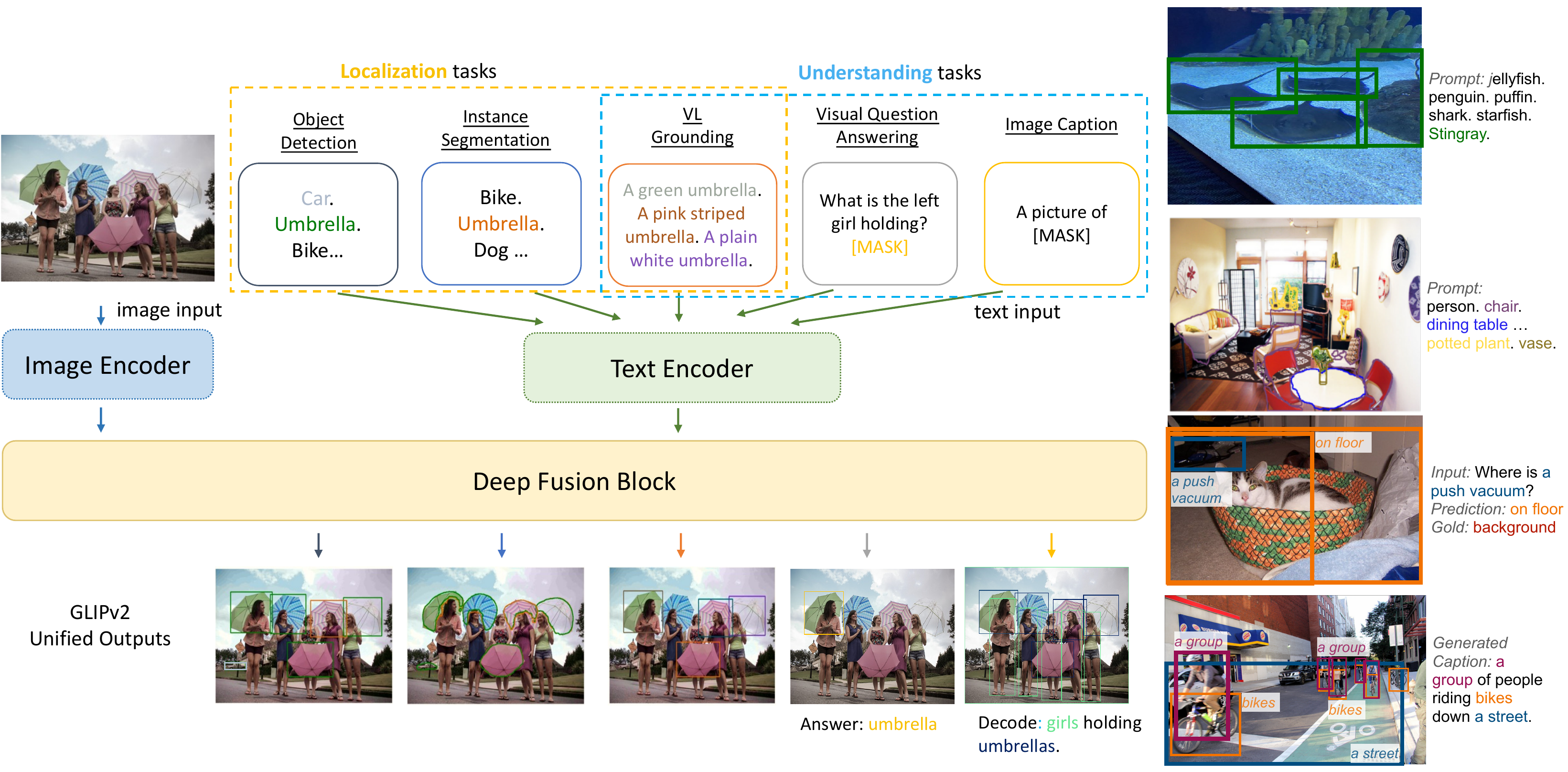}
    \caption{Left: GLIPv2, a pre-trained grounded VL understanding model, unifies various localization and VL understanding tasks. These two kinds of tasks mutually benefit each other, and enables new capabilities such as language-guided detection/segmentation and grounded VQA/captioning. Right: Additional examples from ODinW (detection), LVIS (segmentation), VQA, COCO Captioning.}
    \label{fig:unfied_model}
\end{figure}

In this paper, we identify ``VL grounding'' as a ``meta''-capability for localization and understanding capabilities. 
VL grounding involves not only \emph{understanding} an input sentence but also \emph{localizing} the mentioned entities in the image (see an example in Figure~\ref{fig:unfied_model}). We build a \textbf{grounded VL understanding} model (\our) as a unified model for localization and VL understanding tasks. 

\textbf{Localization + VL understanding = grounded VL understanding}. Localization tasks involve both localization and semantic classification, where classification can be cast as a VL understanding problem using the {\it classification-to-matching} trick (Section \ref{sec:unif_arch}). Therefore, we reformulate localization tasks as VL grounding tasks, in which the language input is a synthesized sentence as the concatenation of category names~\cite{li2022grounded}. Localization data are turned into VL grounding data, accordingly. The massive VL understanding data (image-text pairs) can be easily turned into VL grounding data in a self-training manner~\cite{li2022grounded}. Therefore, \our has a unified pre-training process: all task data are turned into grounding data and \our is pre-trained to perform grounded VL understanding. 

\textbf{A stronger VL grounding task: inter-image region-word contrastive learning}. GLIP~\cite{li2022grounded} proposes the phrase grounding task as its pre-training task, which we argue is an easy task and does not fully utilize data information. For example, in the VL grounding task in Figure~\ref{fig:unfied_model}, the phrase grounding task only requires the model to match a given image region to one of the three phrases in the text input, i.e., ``green, pink striped, or plain white umbrella?''. This 1-in-3 choice is very easy, only requires color understanding, but loses lots of information in this grounding data: the umbrellas are not any other colors, like black, yellow, etc; objects in those regions are umbrellas but not any other categories, like car, bike, etc. From a contrastive learning view, this phrase grounding task only has two negatives. More negatives can be created from this annotation and thus enable stronger contrastive learning. In \our, we introduce the novel inter-image region-word contrastive learning task, which leverages phrases from other sentences in the same batch as potential negatives, as another much stronger VL grounding task. This new region-word contrastive loss enables \our to learn more discriminative region-word features and demonstrates improvements over all downstream tasks.

\textbf{\our achieves mutual benefit between localization and VL understanding}. 1) Experimental results (Table~\ref{tab:one_set_weight}) show that a single \our model (all model weights are shared) achieves near SoTA performance on various localization and understanding tasks. 2) Thanks to semantic-rich annotations from the image-text data, \our shows superior zero-shot and few-shot transfer learning ability to open-world object detection and instance segmentation tasks, evaluated on the LVIS dataset and the "Object Detection in the Wild (ODinW)" benchmark. 3) \our enables language-guided detection and segmentation ability, and achieves new SoTA performance on the Flick30K-entities phrase grounding and PhraseCut referring image segmentation tasks. 4) Inherently a grounding model, \our leads to VL understanding models with strong grounding ability, which are self-explainable and easy to debug. For example, \our, when \our is finetuned on VQA, it can answer questions while localizing mentioned entities (see Figure~\ref{fig:unfied_model} and Section~\ref{sec:visualize}).

\section{Related Work}
\label{sec:related}

\textbf{Localization models.} Traditionally, localization tasks such as object detection and segmentation are single-modality and output bounding boxes or pixel masks \cite{redmon2016you,lin2017focal,he2017mask,dai2016r,ren2015faster,chen2019hybrid,carion2020end}. One challenge of these single-modality models lies in generalization to rare and novel concepts: it is hard to collect localization data that cover many rare categories \cite{gupta2019lvis}. A long line of research focuses on this generalization problem, under the name of zero-shot \cite{bansal2018zero,zhu2019zero,bucher2019zero,zhu2020don}, weakly-supervised \cite{gokberk2014multi,bilen2016weakly,wang2022omni}, or open-vocabulary \cite{zareian2021open,gu2021zero} localization. Built upon MDETR~\cite{kamath2021mdetr} and GLIP~\cite{li2022grounded}, \our converts localization tasks into a grounded vision-language task using the classification-to-matching trick (Section \ref{section:method}). Thus \our can learn from the semantic-rich vision-language data and shows strong performance on open-vocabulary localization tasks. 

\textbf{Vision-language understanding models.} Vision-language (VL) understanding tasks such as VQA~\cite{VQA}, image captioning~\cite{chen2015microsoft}, and image-text retrieval~\cite{karpathy2014deep} involve understanding visual semantics and how they are expressed in natural language. 
Many VL models (e.g., BUTD) \cite{anderson2018bottom,zhang2021vinvl} rely on a pre-trained localization model as their visual encoder; the downside is the pro-longed ``localization->VLP'' pre-training pipeline \cite{lu2019vilbert,tan2019lxmert,chen2019uniter,su2019vl,li2019visualbert,li2019unicoder,zhou2019unified,li2020oscar,li2020unsupervised}. In contrast, \our simplifies the pre-training pipeline and enables {\it grounded} VL understanding for better interpretability (Section \ref{sec:visualize}).

\textbf{Unifying localization and understanding.} \cite{gupta2021towards,hu2021unit,yang2021crossing} made pioneering efforts in unifying localization and understanding. However, localization tasks are still treated as single-modality tasks, while VL tasks involve two modalities. The unification is achieved via straightforward multi-tasking: a low-level visual encoder is shared across tasks and two separate branches are designed for localization and VL understanding. Such unified models do not bring evident mutual benefit and often underperform task-specific models. In contrast, \our identifies grounded VL understanding as a meta-task for localization and understanding. The task unification brings architecture unification: the unified grounded VL understanding model empowers a localization branch with VL capacity, arriving at a unified branch that excels at both tasks.

\textbf{\our vs GLIP.} 
1) GLIP shows that grounded pre-training improves localization. \our further shows grounded pre-training improves VL understanding and thus leads to a unified model for localization and VL understanding. 2) \our introduces the inter-image region-word contrastive loss, which is another and stronger grounding task than the pre-training task in GLIP. The proposed loss can be viewed as a region-word level generalization of the prevalent image-level contrastive learning \cite{li2021align,radford2021learning,yang2022unified}. 3) \our outperforms GLIP on all benchmarks with the same pre-training data.

\section{\our: Unifying Localization and VL Understanding}
\label{section:method}

Based on the reformulation of object detection as a generalized phrase grounding task in GLIP~\cite{li2022grounded}, we unify both localization and VL understanding tasks as grounded vision-language tasks. A grounded vision-language task takes both image and text as inputs, and outputs region-level understanding results (e.g., detection, segmentation) and/or image-level understanding results with associated grounding/localization information (e.g., VQA, image captioning). We will present the unified grounded VL formulation and architecture in Section~\ref{sec:unif_arch}, the pre-training losses in Section~\ref{sec:unif_loss}, and transfer to downstream tasks in Section~\ref{sec:3_3}.

\subsection{A Unified VL Formulation and Architecture}
\label{sec:unif_arch}
At the center of \our's unified formulation is the \textit{classification-to-matching} trick, which reformulates any \textit{task-specific fixed-vocab classification problem as an task-agnostic open-vocabulary vision-language matching} problem. The best example is the reformulation of image classification as image-text matching in CLIP~\cite{radford2021learning}, which enables the model to learn from raw image-text data directly, and achieves strong zero-shot results on open-vocabulary classification tasks. In GLIPv2, we replace every semantic classification linear layer in traditional single-modality vision models with a vision-language matching dot-product layer.  

As illustrated in Figure~\ref{fig:unfied_model}, \our's unified VL architecture is based on the generic architecture we term Architecture $\mathbf{\Pi}$. It consists of a dual encoder, denoted as $\text{Enc}_{V}$ and $\text{Enc}_{L}$, and a fusion encoder, denoted as $\text{Enc}_{VL}$. The model takes an image-text pair $(\text{Img}, \text{Text})$ as input, and extract visual and text features as below:
\begin{equation}\label{eqn:model}
    \mathring{O} = \text{Enc}_{V}(\text{Img}), \quad \mathring{P} = \text{Enc}_{L}(\text{Text}), \quad O, P = \text{Enc}_{VL}(\mathring{O}, \mathring{P}),
\end{equation}
where $(\mathring{O}$, $\mathring{P})$ and $(O, P)$ denote the image/text features {\it before} and {\it after} VL fusion, respectively. 

\textbf{Vision-Language understanding tasks.} Arch $\mathbf{\Pi}$ is the most popular model architecture for VL understanding tasks.
Given the cross-modality fused representations $O$ and $P$, it is straightforward to add lightweight task-specific heads for various VL tasks. For example, GLIPv2 adds a two-layer MLP on top of text features $P$ as the masked language modeling (MLM) head, to perform the MLM pre-training. We provide model details of VQA and image captioning in Section~\ref{sec:3_3}. 

\textbf{(Language-guided) object detection and phrase grounding.} Following GLIP~\cite{li2022grounded}, \our uses the classification-to-matching trick to unify detection and grounding. More specifically, for detection, we simply replace the class logits $S_{\text{cls}} = O W^T$, where $W$ is the weight matrix of the box classifier, with a task-agnostic region-word similarity logits $S_{\text{ground}} = O P^T$, where text features $P$ are label embeddings from a task-agnostic language encoder. As shown in Figure~\ref{fig:unfied_model}, object detection and phrase grounding share the same input/output format and model architecture. See GLIP~\cite{li2022grounded} for more details. Their only difference is the input text format: (1) for object detection, the text input is a string of concatenated candidate object labels; (2) for phrase grounding, the text input is a natural language sentence.  We refer to GLIP~\cite{li2022grounded} for more details. 

\textbf{(Language-guided) instance segmentation and referring image segmentation.}
Given the object detection results, an instance segmentation head is added to classify each pixel within the box into a semantic class. Again, \our uses the classification-to-matching trick to produce a unified instance segmentation head for the standard instance segmentation tasks and the referring image segmentation tasks and leverage both types of data for its pre-training. This classification-to-matching trick can also apply to many other semantic classification heads in single modality CV models (e.g., semantic segmentation) and thus transfers them to language-guided CV models.
    
\subsection{\our Pre-training}
\label{sec:unif_loss}
The GLIPv2 is pre-trained with three pre-training losses: phrase grounding loss $\Lcal_{\text{ground}}$ from a vision-language reformulation of the object detection task, region-word contrastive loss $\Lcal_{\text{inter}}$ from a novel region-word level contrastive learning task, and the standard masked language modeling loss $\Lcal_{\text{mlm}}$ proposed in BERT~\cite{devlin2018bert}.
\begin{equation}\label{eqn:glipv2loss}
    \Lcal_{\text{GLIPv2}} = \underbrace{\Lcal_{\text{loc}} + \Lcal_{\text{intra}}}_{\Lcal_{\text{ground}}} + \Lcal_{\text{inter}} + \Lcal_{\text{mlm}}
\end{equation}

Similar to losses in detection tasks, the grounding loss $\Lcal_{\text{ground}}$ has two parts: the localization loss $\Lcal_{\text{loc}}$ trains localization heads with bounding-box supervision, e.g., RPN loss, box regression loss and/or centerness loss~\cite{tian2019fcos}; the intra-image region-word alignment loss $\Lcal_{\text{intra}}$ is essentially the semantic classification/retrieval loss for each region. 

\textbf{Intra-image region-word alignment loss.} Given one image-text pair $(\text{Img}, \text{Text})$, we obtain the image and text features {\it after} cross-modality fusion $O$ and $P$. The Intra-image region-word alignment loss is computed by
\begin{equation}\label{eqn:intraloss}
    \Lcal_{\text{intra}} = loss(O P^T; T), 
\end{equation}
where $O P^T$ is the similarity score between image regions and word tokens, and $T$ is the target affinity matrix determined by the ground-truth annotations. The loss function $loss$ is typically a cross-entropy loss for two-stage detectors~\cite{ren2015faster} and a focal loss~\cite{lin2017focal} for one-stage detectors.

However, as discussed in Section~\ref{sec:intro}, this intra-image region-word contrastive learning is rather weak in the sense of contrastive learning, due to the limited number of phrases that can one caption can contain. GLIP~\cite{li2022grounded} alleviates this problem by appending a few negative sentences to form a longer text input with more (negative) phrases. However, constrained by the maximal length of text tokens (256 in GLIP and \our), only a few negative sentences can be added and the number of negative phrases remains in the order of 10's. 
This small-negative-example problem also exists in detection data~\cite{li2022grounded} when the input text cannot include all class names in a detection dataset, e.g., Objects365. 

\textbf{Inter-image region-word contrastive loss.}
In GLIPv2, we propose using phrases from other image-text pairs in the same batch as negative examples, which effectively increases the number of negative examples to the order of 1000's, with nearly negligible additional computational cost. 

\begin{figure}[tb!]
    \centering
    \includegraphics[width=0.8\linewidth]{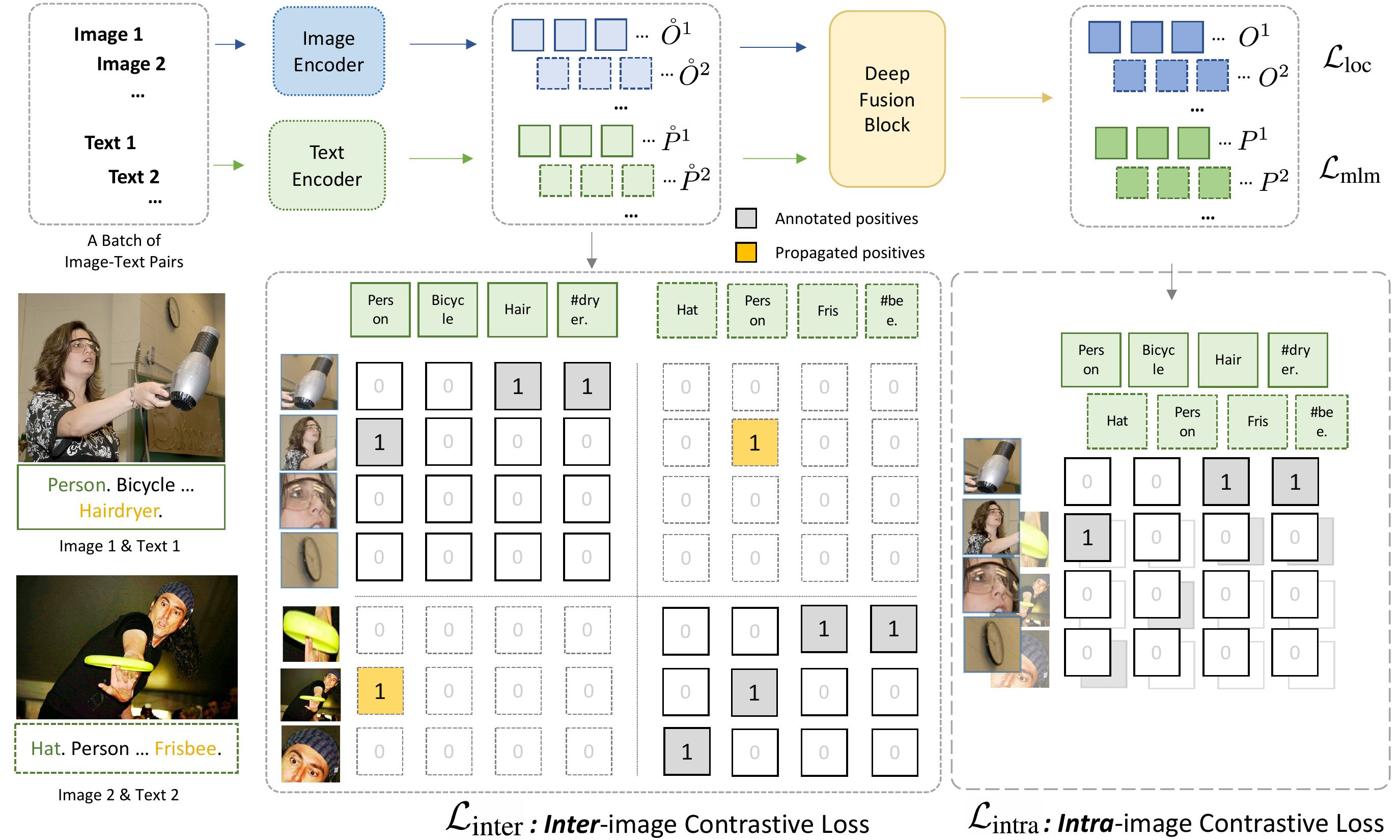}
    \caption{\our pre-training losses: the intra-image alignment loss $\Lcal_{\text{intra}}$ (right) takes features after VL fusion and compute loss over region-word pairs within each image-text pair; the inter-image contrastive loss (left) $\Lcal_{\text{inter}}$ takes features before VL fusion and computes loss over all region-word pairs across a batch of image-text pairs. Label propagation is used to determine the off-diagonal blocks of the $\Lcal_{\text{inter}}$ target matrix~\eqref{eqn:interlogits}. }
    \label{fig:intra_inter_loss}
\end{figure}

As in \eqref{eqn:model}, given a batch of image-text pairs $(\text{Img}^i, \text{Text}^i)_{i=1}^B$ and their ground-truth annotations $(T^i)_{i=1}^B$, the model produces the image and text features {\it before} and {\it after} VL fusion, denoted as $(\mathring{O}^{i}, \mathring{P}^{i})_{i=1}^B$ and $(O^{i}, P^{i})_{i=1}^B$, respectively. Then as illustrated in Figure~\ref{fig:intra_inter_loss} (Left), a batch-wise similarity matrix $S_{\text{ground}}^{\text{batch}}$ and a batch-wise target affinity matrix $T^{\text{batch}}$ are constructed by considering all the image regions and text phrases across this batch. Their $(i,j)$'th blocks are obtained as below:
\begin{equation}\label{eqn:interlogits}
    S_{\text{ground}}^{\text{batch}}[i,j] = \mathring{O}^{i} (\mathring{P}^{j})^T, \quad 
    T^{\text{batch}}[i,j] = \begin{cases}
    T^i,& \text{if } i=j\\
    \text{obtained by label propagation},              & \text{otherwise.}
\end{cases}
\end{equation}
The inter-image region-word contrastive loss is then defined as the standard bi-directional contrastive loss applied on all image regions and phrases in this batch:
\begin{equation}\label{eqn:interloss}
    \Lcal_{\text{inter}} = \text{cross\_entropy\_loss}(S_{\text{ground}}^{\text{batch}}, T^{\text{batch}}, \text{axis} = 0) + \text{cross\_entropy\_loss}(S_{\text{ground}}^{\text{batch}}, T^{\text{batch}}, \text{axis} = 1).
\end{equation}
Compared with that in the inter-image contrastive loss \eqref{eqn:intraloss}, the number of negatives is multiplied by batch size $B$ in this inter-image contrastive loss \eqref{eqn:interloss}.  We elaborate two important details in \eqref{eqn:interlogits}. 
(1) \our uses the image text features $(\mathring{O}^{i}, \mathring{P}^{i})_{i=1}^B$ before VL fusion, {\it not} $(O^{i}, P^{i})_{i=1}^B$ after VL fusion, to compute the batch-wise similarity matrix in the inter-image contrastive loss~\eqref{eqn:interlogits}. Otherwise, the image and text features after VL fusion would have seen the paired information~\eqref{eqn:model}, and thus the model can easily rule out the negatives from misaligned images/texts. (2) We cannot simply assign all regions and texts from unpaired image-text as negative pairs, as done in the standard contrastive loss in CLIP~\cite{radford2021learning}. Instead, we determine the off-diagonal blocks in the target affinity matrix $T^{\text{batch}}$ by {\it label propagation}. For example, as illustrated in Figure~\ref{fig:intra_inter_loss} (Left), if a region is annotated as ``person'', it should be a positive pair with all ``person'' phrases in detection-type texts. We do not propagate positives to grounding-type texts (natural sentences) because phrases in sentences carry contexts that are unique to that image-sentence pair. 

\textbf{Pre-training with both detection and paired-image-text data.} \our pre-training data is in the image-text-target triplet format $(\text{Img}, \text{Text}, T)$, where the target affinity matrix $T$ contains the box-label localization annotations. We also use massive image-text pair data $(\text{Img}, \text{Text})$ to pre-train \our, by generating grounding boxes $\hat{T}$ for phrases in the text with the GLIP pre-trained model from \cite{li2022grounded}. The human-annotated OD/grounding data provides high-fidelity localization supervision, while the massive image-text data greatly improves the concept diversity for \our.

\textbf{Second-stage pre-training of the segmentation head.} \our performs a second-stage pre-training of the language-guided segmentation head on both instance segmentation and image referring segmentation data, while fixing all other parts of the model. 

\subsection{Transfer \our to Localization and VL Tasks}
\label{sec:3_3} 
We introduce two ways to easily transfer \our to various downstream tasks. 
In addition, \our can perform conventional VL tasks (e.g., VQA) along with localization, effectively making every task we consider a ``grounded VL understanding'' task.

\textbf{One model architecture for all.} \our can be transferred to downstream tasks by fine-tuning the model with an (optional) task-specific head. 1) For \textit{detection and segmentation} tasks, no task-specific head is needed as the pre-training architecture can inherently perform detection and segmentation. 2) For \textit{VL} tasks: for VQA, a classification head is added on top of the hidden representation of the start-of-sequence token; for caption generation, we train with a unidirectional language modeling loss, which maximizes the likelihood of the next word given context. We use a unidirectional attention mask and prevent the image part from attending to the text in the fusion layers.

\textbf{One set of weights for all.} There is a growing interest in developing models that can be transferred to various tasks while only changing the least amount of parameters to save training time and storage cost \cite{shin2020autoprompt,lester2021power}. Following GLIP, \our can be transferred to localization tasks in a \textit{zero-shot} or a \textit{prompt-tuning} setting (Section ~\ref{sec:5_2}). One single \our model can serve various tasks, where each task only keeps few or no parameters.
Of particular interest is the prompt tuning setting. For a certain localization task, the text prompt is the same for all input images; thus, we could directly tune $\mathring{P}$, a small prompt embedding matrix, to adapt \our to new tasks. Prompt tuning in a deep-fused model such as \our is different from the conventional linear probing/prompt tuning setting \cite{wang2020frustratingly,radford2021learning,zhou2021learning} in shallow-interacting vision models such as CLIP. The latter can also be viewed as only tuning a small prompt/softmax embedding $P$; however, tuning $P$ only affects the very last layer of the model while the visual representation is still frozen. In contrast, GLIP/\our's visual representation is conditioned on the prompt embedding  $\mathring{P}$; tuning $\mathring{P}$ changes the text, visual, as well as fused embeddings. As a result, prompt tuning in \our is highly effective, often matching the performance of fine-tuning (see Table ~\ref{tab:one_set_weight}). This is in contrast to the common observation in CV that linear probing lags behind fine-tuning by a large gap~\cite{he2021masked}.

\textbf{Grounded VL understanding.} \our also enables grounded VL understanding, where we retain the ability to perform grounding when fine-tuning the model to a downstream VL task. This increases the interpretability of the model. Specifically, we first turn the VL data of the downstream task into grounded VL data using a pre-trained GLIP model. Then we train the model with both the downstream task head and grounding head. For VQA, the model is trained to predict the answer and ground entities in the question as well as the implied entity in the answer; for captioning, the model is trained to predict the next word given the context and ground the current decoded word. 
By tuning localization tasks into a grounded VL task and augmenting VL tasks with grounding ability, we effectively turn every task into a grounded VL understanding task (see examples in Figure \ref{fig:unfied_model}).

\section{Experiments}
\label{sec:exps}





In this section, we show that \our serves as a performant and easy-to-deploy general-purpose vision system. 1) \textbf{One Model Architecture for All} (Section ~\ref{sec:5_1}). \our can be directly fine-tuned to both localization and VL understanding tasks with minimal architecture change. It achieves performance on par with SOTA models with specialized architectures. 2) \textbf{One Model Weight for All} (Section ~\ref{sec:5_2}). \our can be transferred to localization tasks in a zero-shot manner with zero parameter update; with prompt tuning, a single \our model can achieve comparable performance with fully fine-tuned settings on both localization and understanding tasks. 

Following GLIP ~\cite{li2022grounded}, we adopt Swin Transformer~\cite{liu2021swin} as the image encoder $\text{Enc}_{V}$, text transformers~\cite{vaswani2017attention,radford2021learning} as the text encoder $\text{Enc}_{L}$, Dynamic Head~\cite{dai2021dynamic} with language-aware deep fusion~\cite{li2022grounded} as the fusion encoder $\text{Enc}_{VL}$, and Hourglass network~\cite{newell2016stacked} as instance segmentation head feature extractor. We train \our at three scales: \ourT, \ourB, and \ourH. 

\noindent \textbf{\ourT} has the same model config and initialization as GLIP-T: Swin-Tiny and BERT-Base as the dual encoder. The model is pre-trained on the following data: 1) O365, 2) GoldG as in GLIP-T (C), and 3) Cap4M, 4M image-text pairs collected from the web with boxes generated by GLIP-T~\cite{li2022grounded}. 
\noindent\textbf{\ourB/\ourH} are based on Swin-Base/Swin-Huge and the pre-layernorm text transformer~\cite{gao2021clip} as dual encoder, and are initialized from the UniCL~\cite{yang2022unified} checkpoints. We observe much stabler training with GPT-type pre-layernorm transformer~\cite{gao2021clip} than BERT-type post-layernorm transformer. The training data contain: 1) FiveODs (2.78M data) \footnote{Besides O365, it combines with 4 additional OD datasets including COCO~\cite{lin2014microsoft}, OpenImages~\cite{krasin2017openimages}, Visual Genome ~\cite{krishna2017visual}, and ImageNetBoxes~\cite{krizhevsky2012imagenet}}; 2) GoldG as in MDETR~\cite{kamath2021mdetr}; and 3) CC15M+SBU, 16M public image-text data with generated boxes by GLIP-L~\cite{li2022grounded}. \textbf{Segmentation heads} of \our models are pre-trained on COCO, LVIS~\cite{gupta2019lvis} and PhraseCut~\cite{wu2020phrasecut}, with all other model parameters are frozen. 

\textbf{Note} All datasets above were collected by the creators (cited) and consent for any personally identifiable information (PII) was ascertained by the authors where necessary. Due to limited space, we refer to supplementary for details of training recipes and hyper-parameters.


\subsection{One Model Architecture for All}
\label{sec:5_1}

\begin{table}[t]
    \centering
\resizebox{\linewidth}{!}{
\setlength{\tabcolsep}{2.5pt}
    \begin{tabu}{l|c|cccc|cc|cc}
    \toprule
    
    \multirow{2}{*}{Model} & \multirow{2}{*}{Model Type} & COCO-Det & ODinW & LVIS & COCO-Mask & Flickr30K & PhraseCut & VQA & Captioning   \\
          & & (test-dev) & (test) & (minival) & (test-dev) & (test) & (test) & (test-dev / test-std) & (Karpathy-test)\\
    \midrule
    \rowfont{\color{darkgray}}
    Mask R-CNN~\cite{he2017mask} & \multirow{5}{*}{Localization}   & 39.8 & - & 33.3 / - & - / 37.1 & - & - & - & - \\
    DETR~\cite{carion2020end} & & 42.0 & - & 17.8 / - & - & - & - & - & -  \\
    
    DyHead-T~\cite{dai2021dynamic} & 
    & 49.7 & 60.8 & - & - & - & - & - & - \\
    \rowfont{\color{darkgray}}
    DyHead-L~\cite{dai2021dynamic}  &  & 60.3* & -  & - & - & - & - & - & - \\
    
    \midrule
    \rowfont{\color{darkgray}}
    VisualBERT~\cite{li2019visualbert}  &  \multirow{3}{2.5cm}{\centering Understanding} 
        & - & - & - & - & 71.33 & - & 70.8 / 71.0  & - \\
    \rowfont{\color{darkgray}}
    UNITER~\cite{chen2020uniter} & & - & - & - & - & - & - & 73.8 / 74.0 &  -  \\
    \rowfont{\color{darkgray}}
    VinVL~\cite{zhang2021vinvl} & & - & -  & - & - & - & - & \textbf{76.5 / 76.6} & 130.8  \\
    
    \midrule
    GPV~\cite{gupta2021towards} & \multirow{4}{3cm}{\centering Localization \& Understanding} & - & -  & - & - & - & - & 62.5 / - & 102.3 \\
    UniT~\cite{hu2021unit} & & 42.3 & - & -  & - & - & - & 67.6 / - & -  \\
    MDETR~\cite{kamath2021mdetr} &  & - & - & 24.2 / - & - & 84.3 & 53.7 & 70.6 / 70.6 & -  \\
    Unicorn~\cite{yang2021crossing} &  & - & - & - & - & 80.4 & - & 69.2 / 69.4 & 119.1 \\
    
    \midrule
    GLIP-T~\cite{li2022grounded} & \multirow{2}{3cm}{\centering Localization \& Understanding} & 55.2 & 64.9 & - & - & 85.7 & - & - & -  \\
    GLIP-L~\cite{li2022grounded} & & 61.5* & 68.9 & - & - & 87.1 & - & - & -  \\
    
    \midrule
    \midrule
    
     GLIPv2-T (\textbf{Ours}) & \multirow{3}{2cm}{\centering Localization \& Understanding} & 55.5 & 66.5 & 50.6 / 41.4 & 53.5 / 42.0 & 86.5 & 59.4 & 71.6 / 71.8   & 122.1 \\
    GLIPv2-B (\textbf{Ours}) & & 58.8 & 69.4 & 57.3 / 46.2 & 59.0 / 45.8  & 87.5 & \textbf{61.3} & 73.1 / 73.3 &  128.5 \\
    GLIPv2-H (\textbf{Ours}) & & \textbf{60.6 (62.4*)} & \textbf{70.4} & \textbf{59.8 / 48.8} & \textbf{59.8 / 48.9} & \textbf{87.7} & \textbf{61.3} & 74.6 / 74.8 &  \textbf{131.0} \\


    \bottomrule
    \end{tabu}
    }
    \caption{One model architecture results. For COCO-Det test-dev, * indicates multi-scale evaluation. For LVIS, we report the numbers for both \texttt{bbox} and \texttt{segm} on minival to avoid data contamination due to the pre-training. For Flickr30K test, we report the metric under \texttt{R@1}. For COCO-Mask, we also report both \texttt{bbox} and \texttt{segm} on test-dev. }
    \label{tab:one_model_arc}
\end{table}

We compare \our to existing object detection and vision-language pre-training methods on a wide range of tasks. We fine-tune the model on 8 different downstream tasks and report the performance in Table \ref{tab:one_model_arc}. We make the following observations.

\textbf{\our v.s. specialized Localization methods.} 
\our outperforms previous localization models on generalization to both common and rare classes and domains \textit{with a single model architecture and pre-training stage}.
\textit{1) OD on common categories (COCO-Det)}, {\ourT} achieves 5.8 improvement compared to the standard DyHead-T trained on O365 (55.5 v.s. 49.7). {\ourH} reaches 62.4 AP on test-dev, and surpass the performance of the previous SoTA model GLIP-L. 
\textit{2) OD on rare / unseen categories (LVIS)}, {\ourT} outperforms a supervised MDETR on the \texttt{bbox} by a great margin (59.8 v.s. 24.2).
\textit{3) Generalization to diverse real-word tasks (ODinw)}, \ourT (55.5) performs better than original GLIP-T (64.9) on the average of 13 public datasets; {\ourB} outperforms GLIP-L by 0.5 AP. \textit{4) Instance segmentation (COCO-Mask \& PhraseCut)}, for traditional instance segmentation (i.e., COCO-Mask), \ourH outperforms the well-known Mask R-CNN by a great margin on \texttt{segm}. For language-guided segmentation (i.e., PhraseCut), compared to MDETR, \ourT achieves an improvement of 5.7 mask AP.

\textbf{\our v.s. specialized VL Understanding methods.} \our rivals with SoTA specialized models for VL tasks. \textit{1) For VQA}, \our outperforms VisualBERT and UNITER and approaches the previous SoTA model VinVL. \textit{2) For Captioning}, the best \our even surpasses VinVL (VinVL and \our are not trained with CIDEr optimization). 

\textbf{\our v.s. localization and VL models.} Prior works such GPV, UniT and Unicorn have also explored unifying localization and VL models (see a discussion in Section~\ref{sec:related}).  \our outperforms all previous systems on both localization and VL tasks. For the best \ourH, it outperforms the UniT by a great margin (18.3 AP) on COCO object detection tasks. Meanwhile, it also surpasses UniT's performance on VQA by 6.9 points and GPV's peformance on Image Captioning as well.  

\textbf{Takeaway.} Most notably, \our outperforms previous ``unified'' models (GPV, UniT, MDETR, Unicorn) by a large margin. This is the first time that a single model architecture could achieve near SoTA performance on both localization and understanding. In contrast, in prior work, there exists certain trade-off between localization and understanding: models that aim to achieve high understanding performance tend to have lower localization performance (e.g., UNiT's detection performance is limited to the DETR~\cite{carion2020end} architecture), as it is not trivial to merge a SoTA localization branch and a SoTA VL branch into a single model.

\subsection{One Set of Model Parameters for All}
\label{sec:5_2}

\our is pre-trained to perform grounding; thus it can be transferred to various localization tasks with changing zero or few parameters. We evaluate \our under two such settings: 1) direct evaluation, where we transfer the model ``as is'' without any parameter change, and 2) prompt tuning, where only the prompt embedding is tuned for specific tasks (Section \ref{sec:3_3}).

\begin{table}[t]
    \centering
\resizebox{\linewidth}{!}{
    \begin{tabular}{l|cccc|cccccc}
    \toprule
    \multirow{3}{*}{Model} &  \multicolumn{4}{c|}{Direct Evaluation} & \multicolumn{5}{c}{Prompt Tuning} \\
    \cmidrule{2-10}
    & COCO-Mask & ODinW & LVIS-Det & Flickr30K & COCO-Det & ODinW  & LVIS & COCO-Mask & PhraseCut \\
    & (minival) & (test) & (minival) & (minival) & (test-dev) & (test) & (minival) & (test-dev) & (test) \\
    \midrule
    GLIP-T & 46.6/-- & 46.5 & 26.0 & \textcolor{gray}{85.7} & -- & 46.5 & -  & - & - \\
    GLIP-L & 49.8/-- & 52.1 & 37.3 & \textcolor{gray}{87.1} & 58.8 & 67.9 & - & - & - \\
    
    \midrule
    \our-T & \textbf{47.3}/\textcolor{gray}{35.7} & 48.5 & \textbf{29.0} & \textcolor{gray}{86.0} 
        & 53.4 \tiny{\textcolor{red}{(-2.1)}} & 64.8 \tiny{\textcolor{red}{(-1.7)}} & 49.3 / 34.8 \tiny{\textcolor{red}{(-1.3 / -6.6)}} & 53.2 / 41.2 \tiny{\textcolor{red}{(-0.3 / -0.8)}} & 49.4 \\
    
    \our-B & \textcolor{gray}{61.9$^\dagger$}/\textcolor{gray}{43.4} & 54.2 & \textcolor{gray}{48.5}  & \textcolor{gray}{87.2} 
        & 59.0 \tiny{\textcolor{red}{(+0.2)}} & 67.3 \tiny{\textcolor{red}{(-2.1)}} & 56.8 / 41.7 \tiny{\textcolor{red}{(-0.5 / -4.5)}} & 58.8 / 44.9 \tiny{\textcolor{red}{(-0.2 / -0.9)}} & 55.9 \\  
    
    \our-H & \textcolor{gray}{64.1$^\dagger$}/\textcolor{gray}{47.4} & \textbf{55.5} & \textcolor{gray}{50.1} & \textbf{\textcolor{gray}{87.7}} & \textbf{60.2 / 61.9}* \tiny{\textcolor{red}{(-0.4 / -0.5)}} & \textbf{69.1} \tiny{\textcolor{red}{(-1.3)}} & \textbf{59.2 / 43.2} \tiny{\textcolor{red}{(-0.6 / -5.7)}}& \textbf{59.8 / 47.2} \tiny{\textcolor{red}{(-0.0 / -1.7)}} & \textbf{56.1} \\ 
    \bottomrule
    \end{tabular} }
    \caption{One set of weights results v.s. Original GLIP. * indicates multi-scale evaluation. Numbers in \textcolor{red}{red} clearly points out the difference between the prompt tuning and full fine-tuning results (see Table ~\ref{tab:one_model_arc}). Numbers in \textcolor{gray}{gray} mean that they are not in \textit{zero-shot} manner. $\dagger$: these two numbers are artificially high due to some overlap between COCO-minival and VisualGenome-train.}
    \label{tab:one_set_weight}
\end{table}

\begin{figure}[t]
\begin{floatrow}
\ffigbox{
    \centering
  \includegraphics[width=0.85\linewidth]{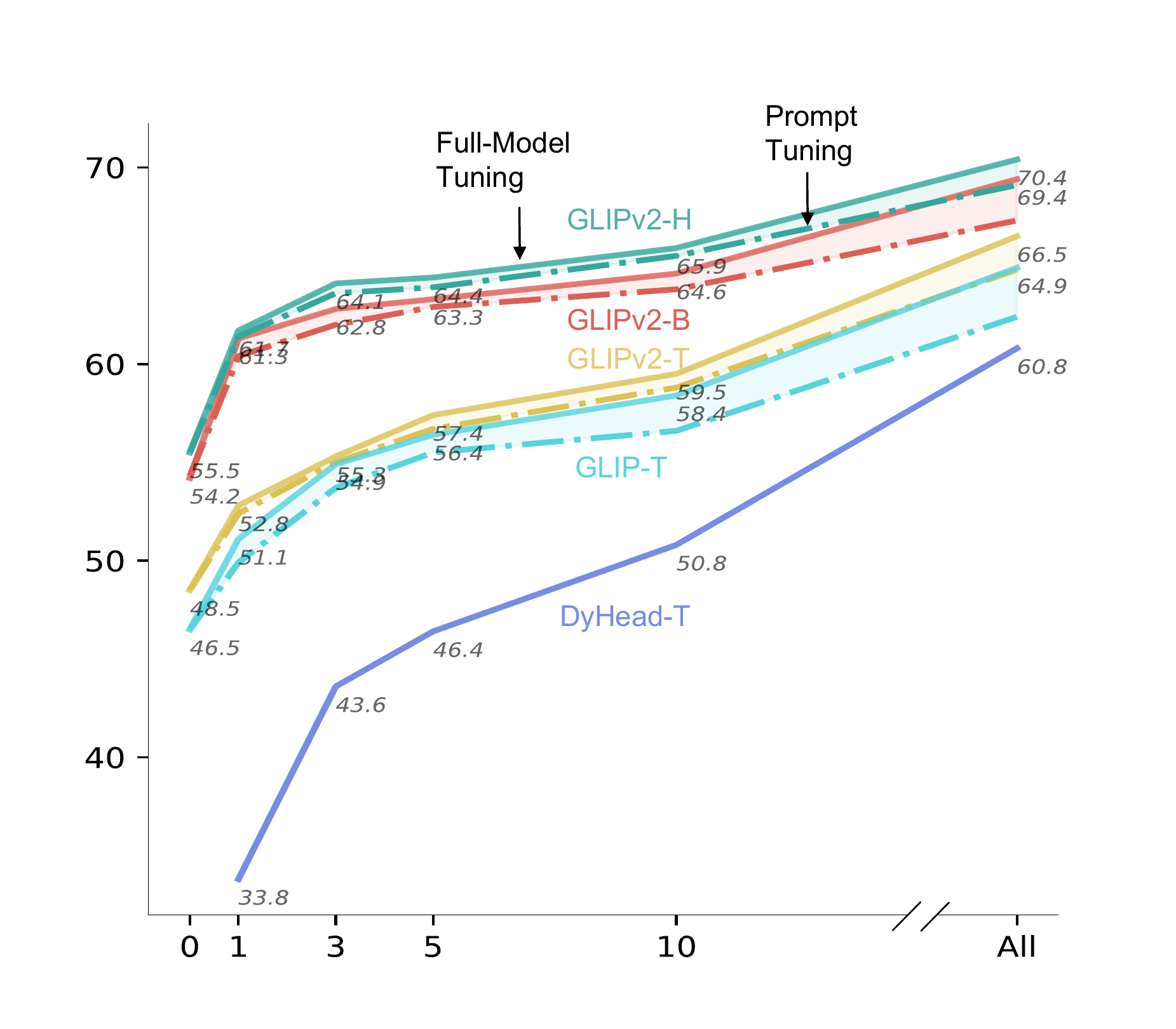}
  \vspace{-10pt}
  }{
    \caption{Data efficiency of \our on ODinW. The X-axis is the amount of task-specific data, from zero-shot to all data. Y-axis is the average AP across 13 datasets.
    }
    \label{fig:tiny_data_efficiency}
    }
\capbtabbox
{\resizebox{\linewidth}{!}{
    \small
    \centering
    
    \begin{tabular}{l|c|cccccc}
    \toprule
    \multirow{2}{*}{Model} & Zero-Shot & \multicolumn{5}{c}{\textcolor{airforceblue}{Prompt Tuning} / \textcolor{amaranth}{Fine Tuning}}\\
    & 0 & 1 & 3 & 5 & 10 & All \\
    \midrule
    \multirow{2}{*}{DyHead-T {\scriptsize{O365}}~\cite{li2022grounded}} & \multirow{2}{*}{-} &  - & - & - & - & - \\  
     &  & \textcolor{amaranth}{33.8} & \textcolor{amaranth}{43.6} & \textcolor{amaranth}{46.4} & \textcolor{amaranth}{50.8} & \textcolor{amaranth}{60.8} \\ 
    \midrule
     \multirow{2}{*}{ $\Lcal_{\text{loc}} + \Lcal_{\text{intra}}$ (GLIP-T)} & 
        \multirow{2}{*}{46.5} 
         & \textcolor{airforceblue}{49.9} & \textcolor{airforceblue}{53.7} & \textcolor{airforceblue}{55.5} & \textcolor{airforceblue}{56.6} & \textcolor{airforceblue}{62.4} \\ 

        & & \textcolor{amaranth}{51.3} & \textcolor{amaranth}{54.9} & \textcolor{amaranth}{56.4} & \textcolor{amaranth}{58.4} & \textcolor{amaranth}{64.9} \\ 
        \midrule
            
        \multirow{2}{*}{$\Lcal_{\text{loc}} + \Lcal_{\text{intra}} + \Lcal_{\text{inter}}$} & 
        \multirow{2}{*}{48.4}  
        & \textcolor{airforceblue}{52.1} & \textcolor{airforceblue}{55.6} & \textcolor{airforceblue}{56.7} & \textcolor{airforceblue}{58.3} & \textcolor{airforceblue}{62.9} \\ 
        
        &    & \textcolor{amaranth}{51.4} & \textcolor{amaranth}{55.3} & \textcolor{amaranth}{56.6} & \textcolor{amaranth}{59.5} & \textcolor{amaranth}{66.3} \\ 
        \midrule
            
         \multirow{2}{*}{$\Lcal_{\text{loc}} + \Lcal_{\text{intra}} + \Lcal_{\text{inter}} + \Lcal_{\text{mlm}}$} & 
        \multirow{2}{*}{48.5}  & \textcolor{airforceblue}{52.4} & \textcolor{airforceblue}{55.6} & \textcolor{airforceblue}{57.4} & \textcolor{airforceblue}{58.8} & \textcolor{airforceblue}{64.8} \\ 
    
        & & \textcolor{amaranth}{52.8} & \textcolor{amaranth}{55.6} & \textcolor{amaranth}{57.4} & \textcolor{amaranth}{59.7} & \textcolor{amaranth}{66.5} \\ 

    \bottomrule
    \end{tabular}
    }}
{\caption{Zero-shot, prompt tuning, and full fine-tuning performance on ODinW. \our models exhibit superior data efficiency.}\label{tab:data_efficiency}}
\end{floatrow}
\end{figure}

\paragraph{Direct evaluation.} 

The pre-trained \our can be directly evaluated on any object detection task (by concatenating the object categories into a text prompt) and visual grounding task without any further tuning. We evaluate the models on four localization tasks: COCO, ODinW, LVIS, and Flickr30, and their results are presented in Table ~\ref{tab:one_set_weight}. Note that for \our-B and \our-H, the training sets of Flick30K and LVIS are present in the pre-training data. Thus, reported numbers on these metrics are not \textit{zero-shot} evaluation (we have marked them \textcolor{gray}{gray}). For all other evaluation results, the models are evaluated in \textit{zero-shot} settings without any further tuning.

\textit{\our can be effortlessly transferred to different localization tasks without further tuning.}  1) For \textit{COCO}, \our-T achieves a zero-shot performance of 47.3 without seeing any COCO training images. This surpasses well-established supervised systems (e.g., Mask R-CNN) and also outperforms GLIP-T by 0.7 AP. 2) For \textit{ODinW}, \our also shows strong zero-shot performance. \ourT (48.5) surpasses the GLIP-T (46.5). Meanwhile, the zero-shot performance of \ourB and \ourH even surpasses the 10-shot tuning performance of DyHead-T (to be introduced in Figure ~\ref{fig:tiny_data_efficiency}). 3) For \textit{LVIS}, \ourT achieves a 3 AP improvement performance compared to the GLIP-T. 4) For \textit{Flickr30K}, \ourB achieves even higher number (87.2) compared to original GLIP-L (87.1). 

\paragraph{Prompt Tuning.} 
Following GLIP, \our supports efficient prompt tuning: the visual representation is heavily conditioned on the text representation due to the deep fusion block (Section ~\ref{sec:3_3}); thus we could fine-tune only the prompt embedding for each task but still maintain high performance. 

\textit{Prompt tuning \our achieves similar performance as full fine-tuning.} 
When comparing the performance of each task in Table ~\ref{tab:one_model_arc} and ~\ref{tab:one_set_weight} at the same time, for \our, prompt tuning performance almost matches the one model architecture results on localization tasks, without changing any of the grounding model parameters. 

\subsection{\our as a Strong Few-Shot Learner}
\label{sec:5_3}

We demonstrate \our's performance on ODinW datasets with respect to different amounts of training data in Figure ~\ref{fig:tiny_data_efficiency}. The performance improvement between \ourT and GLIP-T exhibits more superior data efficiency for prompt tuning. We compare with the SoTA detector DyHead-T, pre-trained on Objects365 in Table ~\ref{tab:data_efficiency}. It can be seen that a zero-shot \ourT (48.5) outperforms a outperforms 5-shot DyHead-T (46.4) while the performance of one-shot \ourH (61.3) surpasses a all-shot fully supervised DyHead-T (60.8).

\subsection{Analysis}
\label{sec:5_4}

\textbf{Pre-training losses} Table ~\ref{tab:ablation_on_tiny} shows the performance of the downstream tasks with different variants of our method. Compared to the GLIP pre-training tasks with only intra-image region-word contrastive loss (Row 3), adding inter-image word-region loss (Row 5) substantially improves the pre-trained model performance across all the object detection tasks (COCO, ODinW, and LVIS) on both zero-shot and fine-tuned manner. Consistent with common observations from most VL understanding methods, adding MLM loss (Row4) benefits for learning the representation for understanding tasks (Flick30k, VQA, and Captioning). Furthermore, using all three losses together at the 1st stage pre-training and doing the 2nd stage pre-training without MLM on OD and GoldG data, \our (Row6) can perform well on both the localization and VL understanding tasks. 

An additional stage of pre-training is applied for small models (\ourT and \ourB) due to limited model capacity. In order to achieve higher performance on both localization and understanding tasks, we find that including all data (even with some noise) and MLM loss in the first stage of pre-training will benefit the model for learning a better representation of both localization and understanding capability. Since the OD tasks require the model with more accurate localization ability, in our 2nd stage of pre-training, we decide to eliminate the MLM loss. The large model (\ourH) does not need this additional stage because it has enough capacity to learn both word-region alignment and MLM together in a single stage.

\textbf{Pre-training data} Table ~\ref{tab:data_scaleup} reports the last checkpoint results on \our when we do the scaling up of pre-training data. As more weak image-text pair data (Cap) is involved in our training, it benefits both standard/in-domain (i.e., COCO, Flickr30K) and large-domain gap (i.e., ODinW, LVIS) tasks. We also show that by adding the inter-image region-word contrastive helps when we are fixing the data at the same scale. For large-domain gap tasks, adding the inter-image region-word contrastive loss will further boost the model to learn better representation. For more detailed scaling-up effects on various tasks under all the checkpoints for GLIP and \our, refer to Appendix. 

Note that the $(\text{Img}, \text{Text}, T)$ data used in \our pre-training can be just human-annotated data (Row1\&2 in Table \ref{tab:data_scaleup}), with which \our pre-training does not involve any pseudo data from a pre-trained grounding/localization model. In order to achieve the best performance, \our uses image-text pair data with pseudo boxes (Cap) from a pre-trained GLIP model (Row3-6 in Table \ref{tab:ablation_on_tiny}), which is trained with the same "grounded VL understanding" task but just with smaller data. 

\begin{table}[t]
    \centering
\resizebox{0.8\linewidth}{!}{
\setlength{\tabcolsep}{2.5pt}
    \begin{tabular}{cl|ccc|ccc}
    \toprule
    Row & Model & COCO & ODinW & LVIS & Flickr30K & VQA & Captioning  \\
    \midrule
    
    1 & No pre-train & --/50.6 & --/60.8 & -- & -- & 64.6 &  111.5  \\
    2 & +  $\Lcal_{\text{mlm}}$ & --/48.5 & --/37.4 & -- & -- & 64.6 &  110.9 \\
    3 & + $\Lcal_{\text{loc}} + \Lcal_{\text{intra}}$ 
        & 46.6/55.2 & 46.5/64.9 & 26.0 & 85.7 & 69.4 & 119.7   \\
    
    4 & + $\Lcal_{\text{loc}} + \Lcal_{\text{intra}} + \Lcal_{\text{mlm}}$  
        & 47.0/55.2 & 47.6/66.2 & 28.5
        & 86.5 & 69.8 & 120.7 \\
    
    5 & + $\Lcal_{\text{loc}} + \Lcal_{\text{intra}} + \Lcal_{\text{inter}}$ 
        & 47.1/55.4 & 48.4/66.3 & 28.6 & 85.8 & 68.7 & 120.4   \\
        
    6 & + $\Lcal_{\text{loc}} + \Lcal_{\text{intra}} + \Lcal_{\text{inter}} + \Lcal_{\text{mlm}}$  
         & 47.3/55.5 & 48.5/66.5 & 29.0 
        & 86.3 & 70.7 &  122.1  \\

        

    \bottomrule
    \end{tabular}
    }
    \caption{Pre-training losses on Tiny-scale model. Involving intra-image region-word alignment loss $\Lcal_{\text{intra}}$, inter-image region-word contrastive loss $\Lcal_{\text{inter}}$ and MLM loss $\Lcal_{\text{mlm}}$ will benefit both localization and understanding tasks.} 
    \label{tab:ablation_on_tiny}
\end{table}

\begin{figure}[t]
\begin{floatrow}
\capbtabbox
{\resizebox{\linewidth}{!}{
    \small
    \centering
    \begin{tabular}{c|l|cccc}
    \toprule
     $\Lcal_{\text{inter}}$ & Pre-train Data & COCO & ODinW & LVIS & Flick30K \\
    \midrule
     \xmark         & O365, GoldG         &   48.06   &   43.14    &  25.6    &   84.36       \\
     \cmark         & O365, GoldG         &   48.59   &    42.64   &   26.9   &   83.90       \\
    \midrule
   \xmark         & O365, GoldG, Cap4M  &   48.21   &    51.35   &  34.2    &    85.56      \\
     \cmark         & O365, GoldG, Cap4M  &   48.79   &    52.70   &   35.0   &     85.50     \\
    \midrule
     \xmark          & O365, GoldG, Cap12M &   48.50   &    49.32   &  35.5    &    85.79      \\
     \cmark         & O365, GoldG, Cap12M &   49.26   &   53.15    &   36.6   &    85.84      \\
    \bottomrule
    \end{tabular}
    }}
{\caption{Pre-train data scale up on Base-scale model. Results are reported at the last checkpoint. See supplementary for results at all checkpoints.}\label{tab:data_scaleup}}

\capbtabbox
{\resizebox{\linewidth}{!}{
\setlength{\tabcolsep}{2.5pt}
\centering
\vspace{-50pt}
\begin{tabular}{l|ccc|ccc}
    \toprule
   \multirow{2}{*}{Model} & \multicolumn{3}{c}{COCO Caption} & \multicolumn{3}{c}{Flickr30K Grounding} \\
     & B4 & CIDEr & SPICE & R@1 & R@5 & R@10 \\
    \midrule
    \ourT & 36.5 & 119.8 & 21.6 & 80.8 & 94.4 & 96.5 \\
    \ourB & 37.4 & 123.0 & 21.9 & 81.0 & 94.5 & 96.5 \\
    \bottomrule
    \end{tabular}}
    }
{\caption{\our can perform captioning and grounding at the same time (a.k.a., grounded VL understanding).}\label{tab:caption_ground}} 
\end{floatrow}
\end{figure}

\textbf{Grounded Vision-Language Understanding}
\label{sec:visualize}
\our can be trained to perform a VL task and grounding at the same time (Section ~\ref{sec:3_3}). We denote such an ability as grounded VL understanding. In Figure ~\ref{fig:unfied_model}, we showcase grounded predictions of \our on VQA and COCO captions. We also conduct quantitative evaluations (Table ~\ref{tab:caption_ground}). The model achieves strong performance for both VL understanding (on COCO Caption) and localization (on Flickr30K Grounding). Such an ability to produce high-level semantic outputs (i.e., answers and captions) and supporting localization results is another appealing trait of \our, as potential users can have a better understanding of the model behaviour. See more detailed analysis and qualitative examples in the Appendix.

\section{Conclusion and Social Impacts}
This paper proposes \our, a unified framework for VL representation learning that serves both localization tasks and VL understanding tasks. We experimentally verify the effectiveness of the unified model and the novel region-word contrastive learning. Compared to existing methods, \our achieves competitive near SoTA performance on various localization and understanding tasks. However, additional analysis of the data and the model is necessary before deploying it in practice since large-scale web data may contain unintended private information, unsuitable images/text, or some bias leakage. Further investigation may be needed for web data due to the above issues.

\section{Acknowledgement}
We thank anonymous reviewers for their comments and
suggestions. Additional thanks go to the Microsoft Research Horizontal AI Team and Microsoft Alexander Multi-modal Team for providing computer resources for large-scale training. The baseline models used in our experiments are based on the open-source code released in the GitHub repository; we acknowledge all the authors who made their code public, which tremendously accelerates our project progress.


\bibliographystyle{splncs04}
\bibliography{egbib}

\newpage

\clearpage
\section*{Appendix}
\appendix

The appendix is organized as follows:

\begin{itemize} 
\item In Section ~\ref{sec:viz}, we provide more visualizations of our model's predictions on various localization and VL understanding tasks. 
\item In Section ~\ref{sec:tasks}, we describe all our evaluated tasks and their dataset in detail. 
\item In Section ~\ref{sec:novelty}, we discuss the difference between our additional inter-image region-word contrastive loss and some other well-known losses that were also applied over a full batch in multiple works. 
\item In Section ~\ref{sec:recipes}, we introduce the training details and hyperparameters used in Section ~\ref{sec:exps} in the main paper. 
\item Section ~\ref{sec:unicl_bert}, we analyze the effect of using different language encoder and their pre-trained weights in our models.
\item In Section ~\ref{sec:pre_data}, we provide more results for all the checkpoints of adding pre-training data (refer to Section 4 in the main paper).  
\item In Section ~\ref{sec:grounded_caption}, we provide a detailed analysis of the experiments of grounded captioning (mentioned in Section ~\ref{sec:exps} in the main paper). 
\item In Section ~\ref{sec:inference_speed}, we give out a comparison for the model's inference speed. 
\item In Section ~\ref{sec:figure_source}, we clearly provide the original sources of the images that are used in our paper. 
\item In Section ~\ref{sec:all_results}, we present per-dataset results for all experiments in ODinW. 

\end{itemize}

\begin{figure}[ht]
    \centering
    \includegraphics[width=0.85\linewidth]{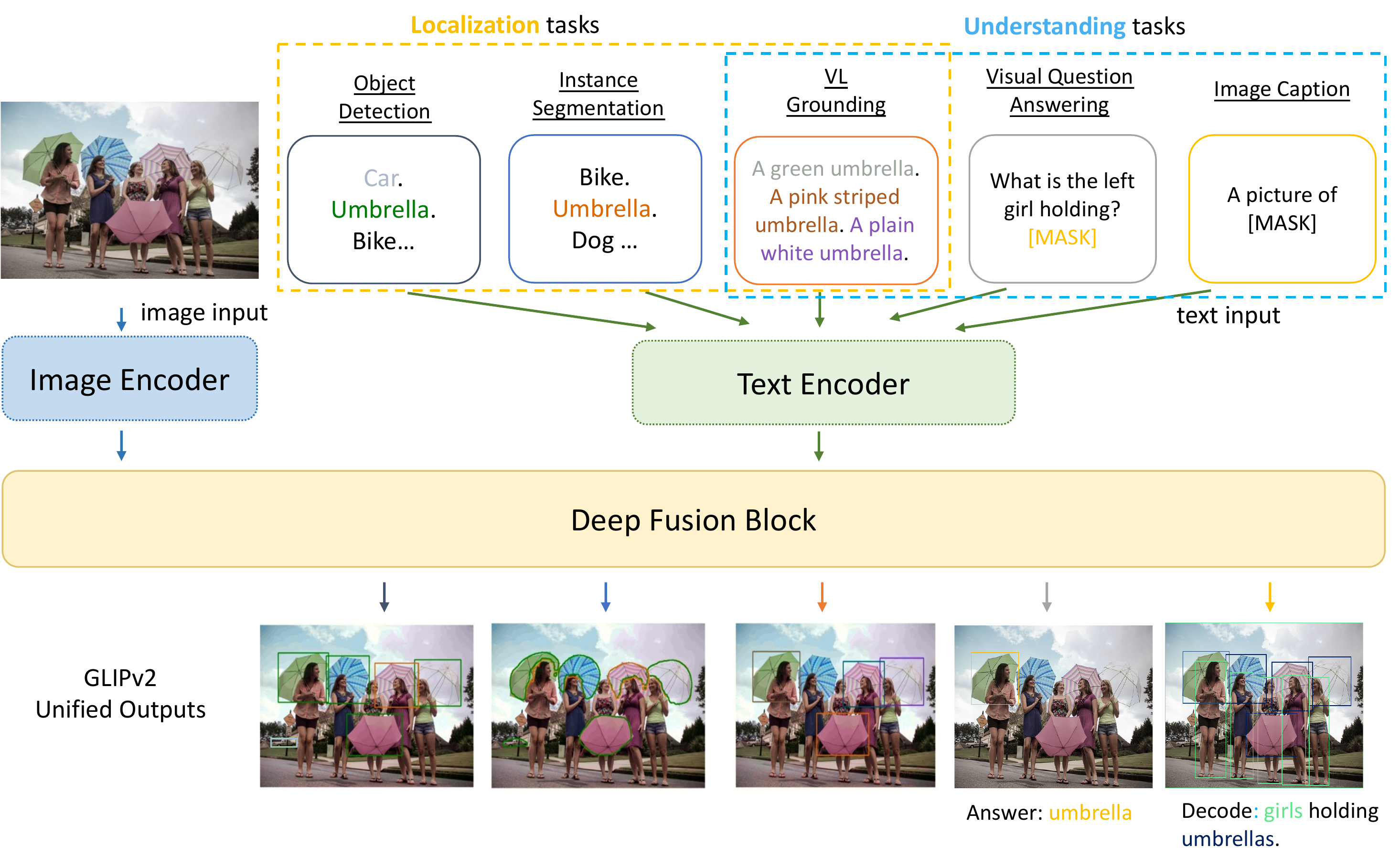}
    \caption{GLIPv2, a pre-trained grounded VL understanding model, unifies various localization and VL understanding tasks. These two kinds of tasks mutually benefit each other and enable new capabilities such as language-guided detection/segmentation and grounded VQA/captioning.}
    \label{fig:big_unfied_model}
\end{figure}

\section{Visualization}
\label{sec:viz}
\begin{figure}[h]
\centering
\includegraphics[width=0.95\textwidth]{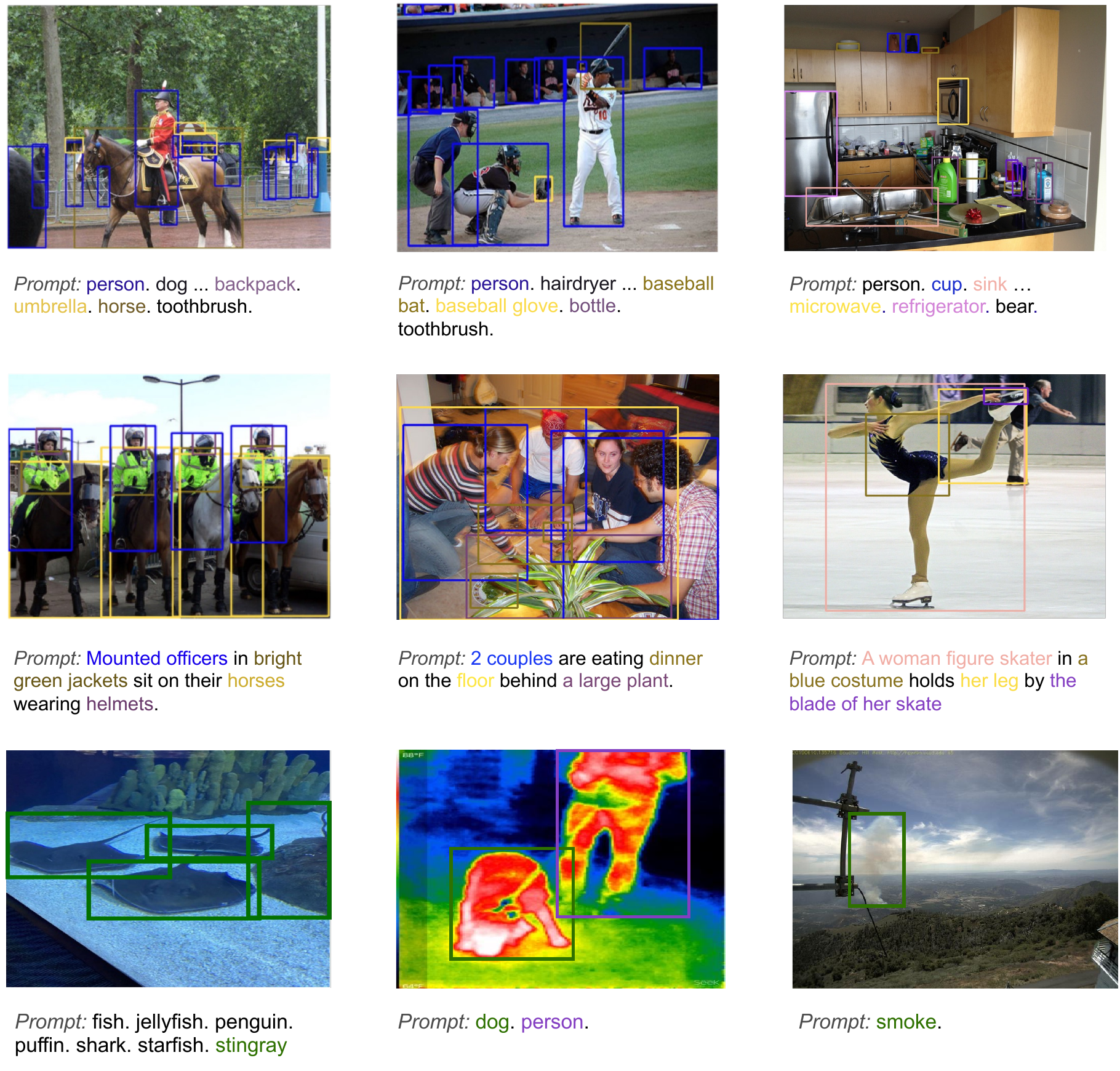} \\
\caption{ Visualization for OD / Grounding. Row 1: Object Detection on COCO. Row 2: Phrase Grounding on Flickr30K. Row 3: Object Detection on ODinW.}
\label{fig:examples_1}
\end{figure}

\begin{figure}[h]
\centering
\includegraphics[width=0.95\textwidth]{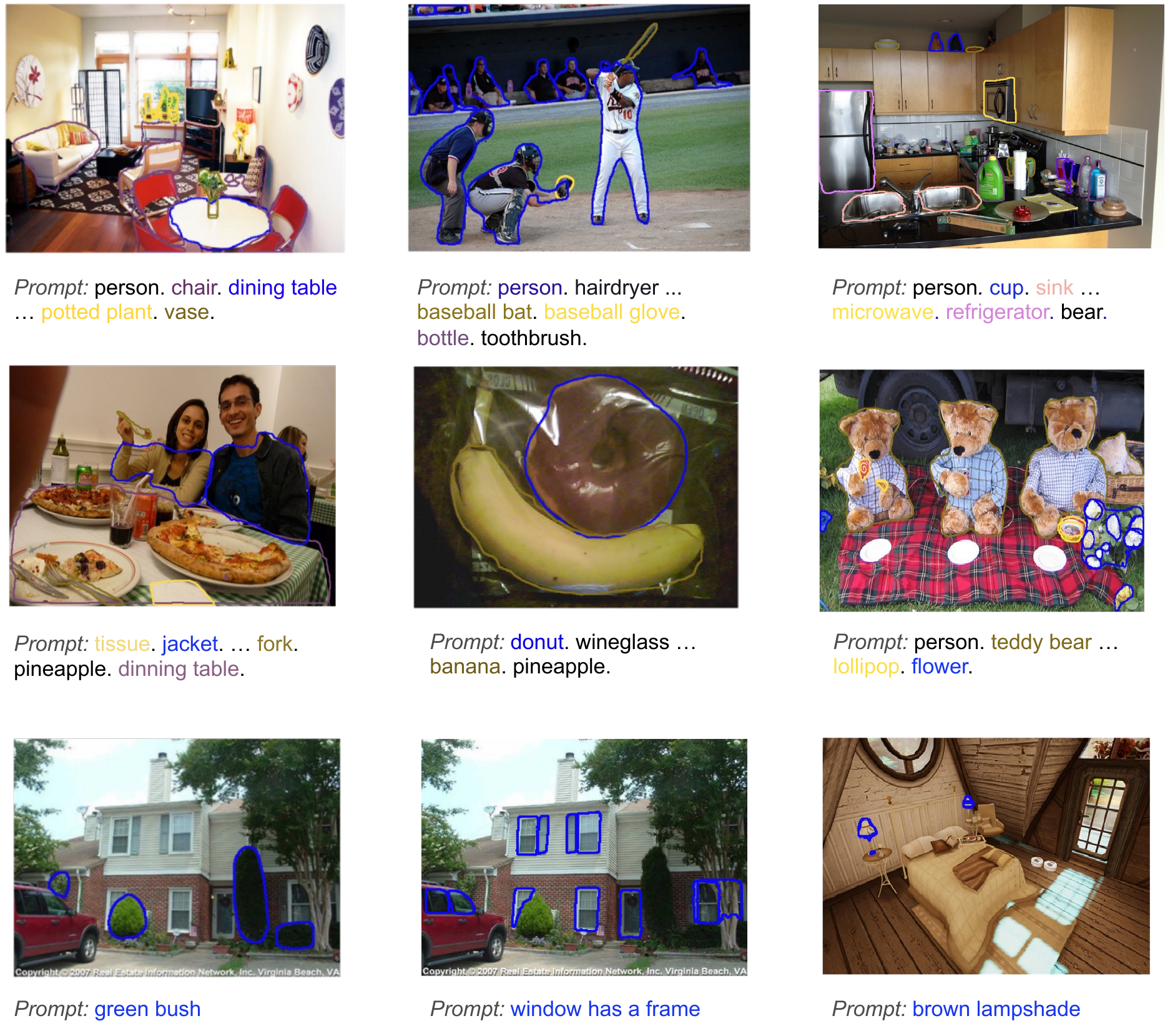}
\caption{ Visualization for Instance / Referring Image Segmentation. Row 1: Instance Segmentation on COCO Mask. Row 2: Instance Segmentation on LVIS. Row 3: Referring Image Segmentation on PhraseCut.}
\label{fig:examples_2}
\end{figure}

\begin{figure}[h]
\centering
\includegraphics[width=0.95\textwidth]{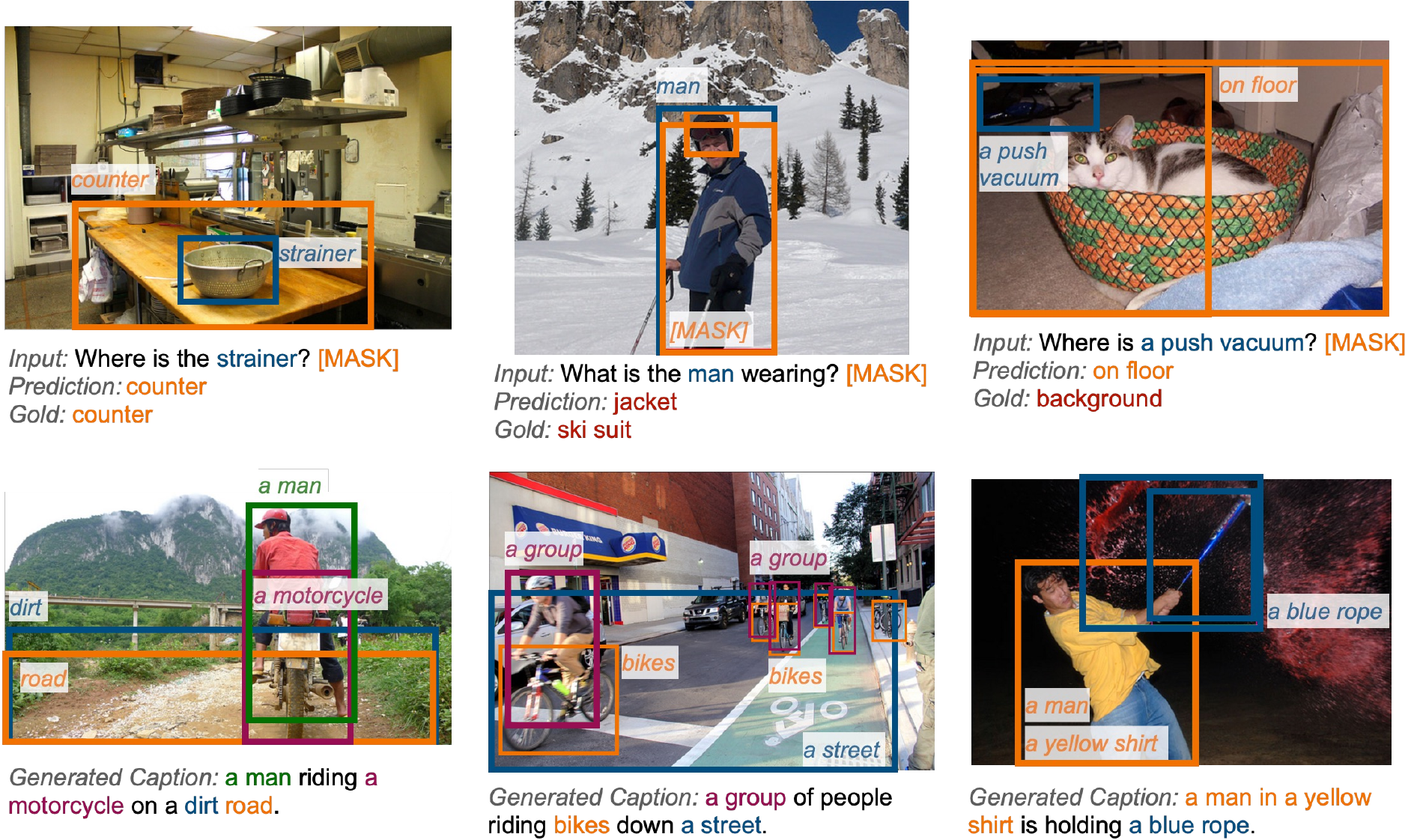}
\caption{Visualization for Grounded VL Understanding. Row 1: Grounded VQA predictions (The model is given the input question and a placeholder token ``[MASK]'' for the answer. The model can ground not only entities in the question but also the implied answer entity). Row 2: Grounded captioning on COCO (The model can generate high-quality captions and, in the meantime, provide localization results.}
\label{fig:examples_3}
\end{figure}

We provide a clearer illustration of \our in Figure ~\ref{fig:big_unfied_model}, which elegantly unifies various localization (object detection, instance segmentation) and VL understanding (phrase grounding, VQA and captioning) tasks. More visualizations of the predictions under various tasks from \our are also provided to indicate the model's strength and capability. Please refer to Figure ~\ref{fig:examples_1} for OD / Grounding, Figure ~\ref{fig:examples_2} for Instance / Referring Image Segmentation, and Figure ~\ref{fig:examples_3} for Grounded VL Understanding. 

\section{Tasks and dataset descriptions}
\label{sec:tasks}
\subsection{(Language-guided) object detection and phrase grounding}

\textbf{COCO.}~\cite{caesar2018coco} The Microsoft Common Objects in Context dataset is a medium-scale object detection dataset. It has about 900k bounding box annotations for 80 object categories, with about 7.3 annotations per image. It is one of the most used object detection datasets, and its images are often used within other datasets (including VG and LVIS).

\textbf{Flickr30k-entities.}~\cite{plummer2015flickr30k} Given one or more phrases, which may be interrelated, the phrase grounding task is to provide a set of bounding boxes for each given phrase. We use the Flickr30k-entities dataset for this task, with the train/val/test splits as provided by \cite{li2022grounded} and evaluate our performance in terms of Recall. Flickr30K is included in the gold grounding data so we directly evaluate the models after pre-training as in MDETR \cite{kamath2021mdetr}. We predict use any-box protocol specified in MDETR.

\textbf{ODinW.} We use 13 datasets from Roboflow\footnote{\url{https://public.roboflow.com/object-detection}}. Roboflow hosts over 30 datasets, and we exclude datasets that are too challenging (e.g., detecting different kinds of chess pieces) or impossible to solve without specific domain knowledge (e.g., understanding sign language). We provide the details of the 13 datasets we use in Table \ref{table:odinw_dataset}. We include the PASCAL VOC 2012 dataset as a reference dataset, as public baselines have been established on this dataset. For PascalVOC, we follow the convention and report on the validation set. For Pistols, there are no official validation or test sets so we split the dataset ourselves. 

\begin{table}[tb!]
\caption{13 ODinW dataset statistics. We summarize the objects of interest for each dataset and report the image number of each split. }
\label{table:odinw_dataset}
\begin{center}
\resizebox{\linewidth}{!}{
\begin{tabular}{l@{\hskip9pt} | 
c@{\hskip9pt}|c@{\hskip9pt}|
c@{\hskip9pt}  
c@{\hskip9pt} c@{\hskip9pt}c@{\hskip9pt}
c@{\hskip9pt}c@{\hskip9pt}c@{\hskip9pt}  c@{\hskip9pt}
c@{\hskip9pt}c@{\hskip9pt}c@{\hskip9pt}c@{\hskip9pt}c@{\hskip9pt}c}
\toprule

Dataset & Objects of Interest & Train/Val/Test & URL \\
\midrule
PascalVOC & Common objects (PascalVOC 2012) & 13690/3422/- & \url{https://public.roboflow.com/object-detection/pascal-voc-2012} \\
AerialDrone & Boats, cars, etc. from drone images & 52/15/7 &  \tiny{\url{https://public.roboflow.com/object-detection/aerial-maritime}}\\
Aquarium & Penguins, starfish, etc. in an aquarium & 448/127/63 & \tiny{\url{https://public.roboflow.com/object-detection/aquarium}} \\
Rabbits & Cottontail rabbits & 1980/19/10 & \tiny{\url{https://public.roboflow.com/object-detection/cottontail-rabbits-video-dataset}} \\
EgoHands & Hands in ego-centric images & 3840/480/480 & \tiny{\url{https://public.roboflow.com/object-detection/hands}} \\
Mushrooms & Two kinds of mushrooms & 41/5/5 &  \url{https://public.roboflow.com/object-detection/na-mushrooms}\\
Packages & Delivery packages & 19/4/3 & \url{https://public.roboflow.com/object-detection/packages-dataset} \\

Raccoon & Raccoon & 150/29/17 & \url{https://public.roboflow.com/object-detection/raccoon} \\

Shellfish & Shrimp, lobster, and crab & 406/116/58 & \url{https://public.roboflow.com/object-detection/shellfish-openimages} \\

Vehicles & Car, bus, motorcycle, truck, and ambulance & 878/250/126 & \url{https://public.roboflow.com/object-detection/vehicles-openimages} \\

Pistols & Pistol & 2377/297/297 & \url{https://public.roboflow.com/object-detection/pistols/1}\\

Pothole & Potholes on the road & 465/133/67 &  \url{https://public.roboflow.com/object-detection/pothole} \\

Thermal & Dogs and people in thermal images & 142/41/20 & \url{https://public.roboflow.com/object-detection/thermal-dogs-and-people} \\

\bottomrule
\end{tabular}
}
\end{center}
\end{table}

\subsection{(Language-guided) instance segmentation and referring image segmentation}

\textbf{LVIS.}~\cite{gupta2019lvis} The Large Vocabulary Instance Segmentation dataset has over a thousand object categories, following a long-tail distribution with some categories having only a few examples. Similar to VG, LVIS uses the same images as in COCO, re-annotated with more object categories. In contrast to COCO, LVIS is a federated dataset, which means that only a subset of categories is annotated in each image. Annotations, therefore, include positive and negative object labels for objects that are present and categories that are not present, respectively. In addition, LVIS categories are not pairwise disjoint, such that the same object can belong to several categories.

\textbf{PhraseCut.}~\cite{wu2020phrasecut} Besides object detection, we show that our GLIPv2 can be extended to perform segmentation by evaluating the referring expression segmentation task of the recent PhraseCut\cite{wu2020phrasecut} which consists of images from VG, annotated
with segmentation masks for each referring expression. These expressions comprise a wide vocabulary of objects, attributes and relations, making it a challenging benchmark. Contrary to other referring expression segmentation datasets, in PhraseCut the expression may refer to several objects and the model is expected to find all the corresponding instances.  

\subsection{VQA and image captioning}
\textbf{VQA.}~\cite{goyal2017making} requires the model to predict an answer given an image and a question. We conduct experiments on the VQA2.0 dataset, which is constructed using images from COCO.  It contains 83k images for training, 41k for validation, and 81k for testing. We treat VQA as a classification problem with an answer set of 3,129 candidates following the common practice of this task. For our best models, we report test-dev and test-std scores by submitting to the official evaluation server.\footnote{\url{https://eval.ai/challenge/830/overview}}

\textbf{COCO image captioning.}~\cite{chen2015microsoft} The goal of image captioning is to generate a natural language description given an input image. We evaluate \our on COCO Captioning dataset and report BLEU-4, CIDEr, and SPICE scores on the Karparthy test split.

\section{Difference between inter-image region-word contrastive loss with other "region-word" losses.}
\label{sec:novelty}
As far as we know, up to the deadline (05/19/2022) for NeurIPS submission, there are only three published papers (VILD \cite{gu2021open}, RegionCLIP \cite{zhong2022regionclip}, and X-VLM \cite{zeng2021multi}) that have the flavor of "region-word" loss applied over full batch. We discuss the difference between our work and the three aforementioned works in the following: 
\begin{enumerate}
    \item All these three works use ``region-sentence" loss, i.e., the similarity between a region feature and the [CLS] token of a sentence, instead of true "region-word" loss used in \our. As a result, none of these three works made use of the phrase grounding data, which may contain multiple entities in one sentence during their training. It is the most important point in \our to use phrase grounding data and pseudo grounding data to train a unified grounded VL understanding model. 
    \item \our has carefully designed the positive label propagation in our inter-image region-word contrastive loss to mitigate the wrong assumption that "every unpaired region-word pair is negative". As far as we know, no previous work has mentioned this mechanism of positive label propagation before.
    \item There are some other differences. For example, in VILD, its ``region-sentence loss" is actually not a contrastive loss over full-batch but a classification loss over a fixed vocabulary per sample (see the definition of $L_{ViLD-text})$.
\end{enumerate}

Upon all three points above, we believe that our inter-image region-word contrastive loss is novel and has a significant difference from previous works. 

\section{Training details and hyperparamters}
\label{sec:recipes}

\begin{table}[bt!]
\resizebox{\linewidth}{!}{
\begin{tabular}{l|c|c|ccc}
\toprule
\multirow{2}{*}{Model} & \multirow{2}{*}{Image} & \multirow{2}{*}{Text} & \multicolumn{3}{c}{Pre-Train Data} \\
 &  &  & Detection & Grounding & Caption \\
 \toprule
GLIPv2-T & Swin-T & BERT-Base & O365 & GoldG (no COCO) & Cap4M \\
GLIPv2-B & Swin-B & CLIP & O365, COCO, OpenImages, VG, ImageNetBoxes & GoldG & CC15M+ SBU \\
GLIPv2-H & CoSwin-H \cite{yuan2021florence} & CLIP & O365, COCO, OpenImages, VG, ImageNetBoxes & GoldG & CC15M+SBU\\
\midrule
Mask Head & -- & -- & LVIS, COCO & PhraseCut & -- \\
\bottomrule
\end{tabular}
}
\caption{A detailed list of \our model variants}
\label{tab:pretrain_setup}
\end{table}


\subsection{Pre-training} 
\textbf{Pre-training data.} There are three different types of data in pre-training 1) detection data 2) grounding data 3) caption data, as shown in Table ~\ref{tab:pretrain_setup}. The detection data includes Object365 ~\cite{shao2019objects365}, COCO~\cite{caesar2018coco}, OpenImages~\cite{krasin2017openimages}, Visual Genome~\cite{krishna2017visual}, and ImageNetBoxes~\cite{imagenet}. The grounding data includes GoldG, 0.8M human-annotated gold grounding data curated by MDETR \cite{kamath2021mdetr} combining Flick30K, VG Caption, and GQA \cite{hudson2019gqa}. The Cap4M is a 4M image-text pairs collected from the web with boxes generated by GLIP-T(C) in \cite{li2022grounded}, and CC (Conceptual Captions) + SBU (with 1M data). 

\textbf{Implementation details.} In Section ~\ref{sec:exps} in the main paper, we introduced GLIPv2-T, GLIPv2-B, GLIPv2-H, and we introduce the implementation details in the following.

\begin{figure}[bt!]
    \centering
    \includegraphics[width=\linewidth]{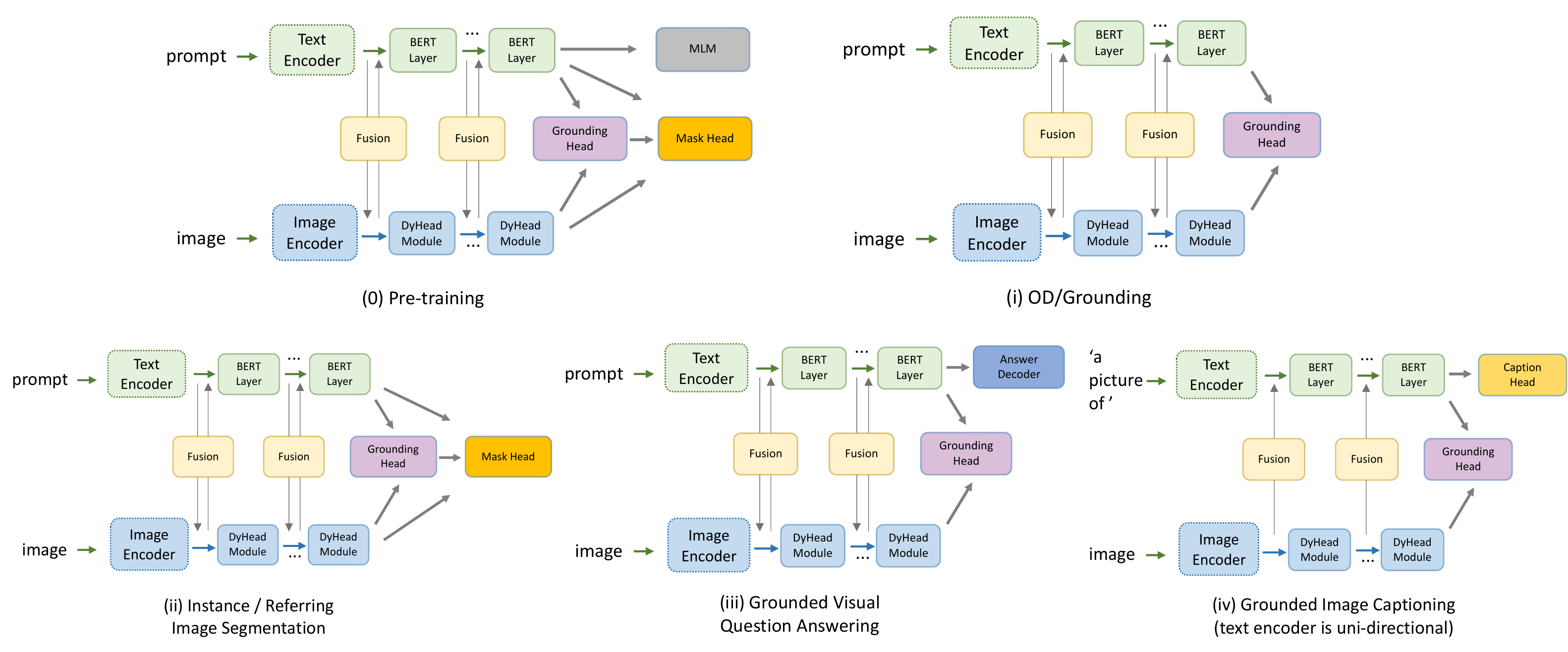}
    \caption{The model architecture for pre-training (0), and downstream tasks (i) OD / Grounding (ii) Instance / Referring Image Segmentation (iii) Grounded Visual Question Answering (iv) Grounded Image Captioning.}
    \label{fig:downstream_arch}
\end{figure}

We pre-train \ourT based on Swin-Tiny models with 32 GPUs and a batch size of 64. We use a base learning rate of $1\times10^{-5}$ for the language backbone (BERT-Base) and $1\times10^{-4}$ for all other parameters. The learning rate is stepped down by a factor of 0.1 at the 67\% and 89\% of the total 330,000 training steps. We decay the learning rate when the zero-shot performance on COCO saturates. The max input length is 256 tokens for all models. To optimize the results for object detection, we continue pre-training without the MLM loss for another 300,000 steps. 

We pre-train \ourB based on Swin-Base models with 64 GPUs and a batch size of 64. We use a base learning rate of $1\times10^{-4}$ for all parameters, including the language backbone (CLIP-type pre-layernorm transformer). The learning rate is stepped down by a factor of 0.1 at the 67\% and 89\% of the total 1 million training steps. We decay the learning rate when the zero-shot performance on COCO saturates. The max input length is 256 tokens for all models. To optimize the results for object detection, we continue pre-training without the MLM loss for another 500,000 steps. 

We pre-train \ourH based on the CoSwin-Huge model from Florence~\cite{yuan2021florence} with 64 GPUs and a batch size of 64. We use a base learning rate of $1\times10^{-4}$ for all parameters, including the language backbone (CLIP-type pre-layernorm transformer). The learning rate is stepped down by a factor of 0.1 at the 67\% and 89\% of the total 1 million training steps. We decay the learning rate when the zero-shot performance on COCO saturates. The max input length is 256 tokens for all models. We found that there is \textbf{no} need to continue pre-training without MLM loss for the huge model.

Mask heads of \ourT, \ourB and \ourH are pre-trained COCO, LVIS and PhraseCut, while freezing all the other model parameters. This mask head pre-training uses batch size 64, and goes through COCO for 24 epochs, LVIS for 24 epochs, and PhraseCut for 8 epochs, respectively. \our uses Hourglass network~\cite{newell2016stacked} as instance segmentation head feature extractor, and utilizes the "classification-to-matching" trick to change the instance segmentation head linear prediction layer (outputs $K$-dimensional logits on each pixel) to a dot product layer between pixel visual features and the word features after VL fusion. \ourT and \ourB use a very basic Hourglass network for segmentation head feature extractor: only 1 scale and 1 layer, with hidden dimension 256. \ourH uses a larger Hourglass network for segmentation head feature extractor: 2 scales and 4 layers, with hidden dimension 384.

\subsection{Downstream tasks}

\textbf{OD / Grounding.} When fine-tuning on COCO, we use a base learning rate of $1\times10^{-5}$ and 24 training epochs for the pre-trained \ourT model, and a base learning rate of $5\times10^{-6}$ and 5 training epochs for the pre-trained \ourB and \ourH models.

For direct evaluation on LVIS, since LVIS has over 1,200 categories and they cannot be fit into one text prompt, so we segment them into multiple chunks, fitting 40 categories into one prompt and query the model multiple times with the different prompts. We find that models tend to overfit on LVIS during the course of pre-training so we closely monitor the performance on minival for all models and report the results with the best checkpoints in Table 2 in the main paper.

For direct evaluation on Flickr30K, models may also overfit during the course of pre-training so we monitor the performance on the validation set for all models and report the results with the best checkpoints in Table 2 in the main paper.

\textbf{Instance segmentation / Referring Image Segmentation.} Given the pre-trained model with pre-trained mask head, we simply fine-tune the \textbf{entire} network to get the task-specific fine-tuned models. 

For fine-tuning on COCO instance segmentation, we use a base learning rate of $1\times10^{-5}$ and 24 training epochs for the pre-trained \ourT model, and a base learning rate of $5\times10^{-6}$ and 5 training epochs for the pre-trained \ourB and \ourH models.

For fine-tuning on LVIS instance segmentation, we use a base learning rate of $1\times10^{-5}$ and 24 training epochs for the pre-trained \ourT model, and a base learning rate of $5\times10^{-6}$ and 5 training epochs for the pre-trained \ourB and \ourH models.

For fine-tuning on PhraseCut Referring Image segmentation, we use a base learning rate of $1\times10^{-5}$ and 12 training epochs for the pre-trained \ourT model, and a base learning rate of $5\times10^{-6}$ and 3 training epochs for the pre-trained \ourB and \ourH models.

\textbf{(Grounded) VQA.} To fine-tune GLIPv2 for VQA, we feed the image and question into the model and then take the output feature sequence $P$ from the language side (after the VL fusion) and apply a `attention pooling' layer to obtain a feature vector $P_{vqa}$. More specifically, the attention pooling layer applies a linear layer followed by softmax to obtain normalized scaler weights, and then these weights are used to compute a weighted sum to produce the feature vector $p_{vqa}$. This feature vector is then fed to a 2-layer MLP with GeLU activation~\cite{hendrycks2016gaussian} and a final linear layer to obtain the logits for the 3129-way classification.\footnote{We experimented simpler pooling methods such as average pooling and \texttt{[CLS]} pooling~\cite{devlin2018bert} in the early experiments and found the attention pooling described above works better.} Following standard practice~\cite{teney2018tips}, we use binary cross entropy loss to take account of different answers from multiple human annotators.
Following VinVL~\cite{Zhang_2021_CVPR}, we train on the combination of train2014 + val2014 splits of the VQAv2 dataset, except for the reserved 2k dev split.\footnote{2000 images sampled from the val2014 split (and their corresponding question-answer pairs).}. For the ablation studies we report the accuracy on this 2k dev split. 

Other than the conventional VQA setting, we also experimented a new `grounded VQA' setup, which the model is required to not only predict the answer, but also ground the objects (predict bounding boxes in the image) mentioned in the question and answer text, see Figure ~\ref{fig:downstream_arch}(iii). Note that the language input is the question appended by a \texttt{[MASK]} token, and this \texttt{[MASK]} token should ground to the object if the answer is indeed an object in the image. The total training loss is summing the grounding loss (intra-image region-word contrastive loss) and the VQA loss described previously.

\textbf{(Grounded) Image Captioning.} We fine-tune the pre-trained model on COCO Caption ``Karpathy'' training split. The training objective is uni-directional Language Modeling (LM), which maximizes the likelihood of the next word at each position given the image and the text sequence before it. To enable autoregressive generation, we use uni-directional attention mask for the text part, and prevent the image part from attending to the text part in the fusion layers. Although the training objective (LM) is different from that in pre-training (i.e., bi-directional MLM), we directly fine-tune the model for image captioning to evaluate its capability of generalizing to VL generation tasks. Our model is trained with cross entropy loss only, without using CIDEr optimization. 

For grounded image captioning (Figure ~\ref{fig:downstream_arch}), we add the grounding loss (intra-image region-word contrastive loss) in training, which is calculated in the same way as in pre-training. We use Flickr30K training split for this task. During inference, for each predicted text token, we get its dot product logits with all the region representations and choose the maximum as the associated bounding box.

\section{Analysis on the effect of different language encoders and pre-trained weights}
\label{sec:unicl_bert}

\begin{figure}[tb!]
\centering
\includegraphics[width=0.8\textwidth]{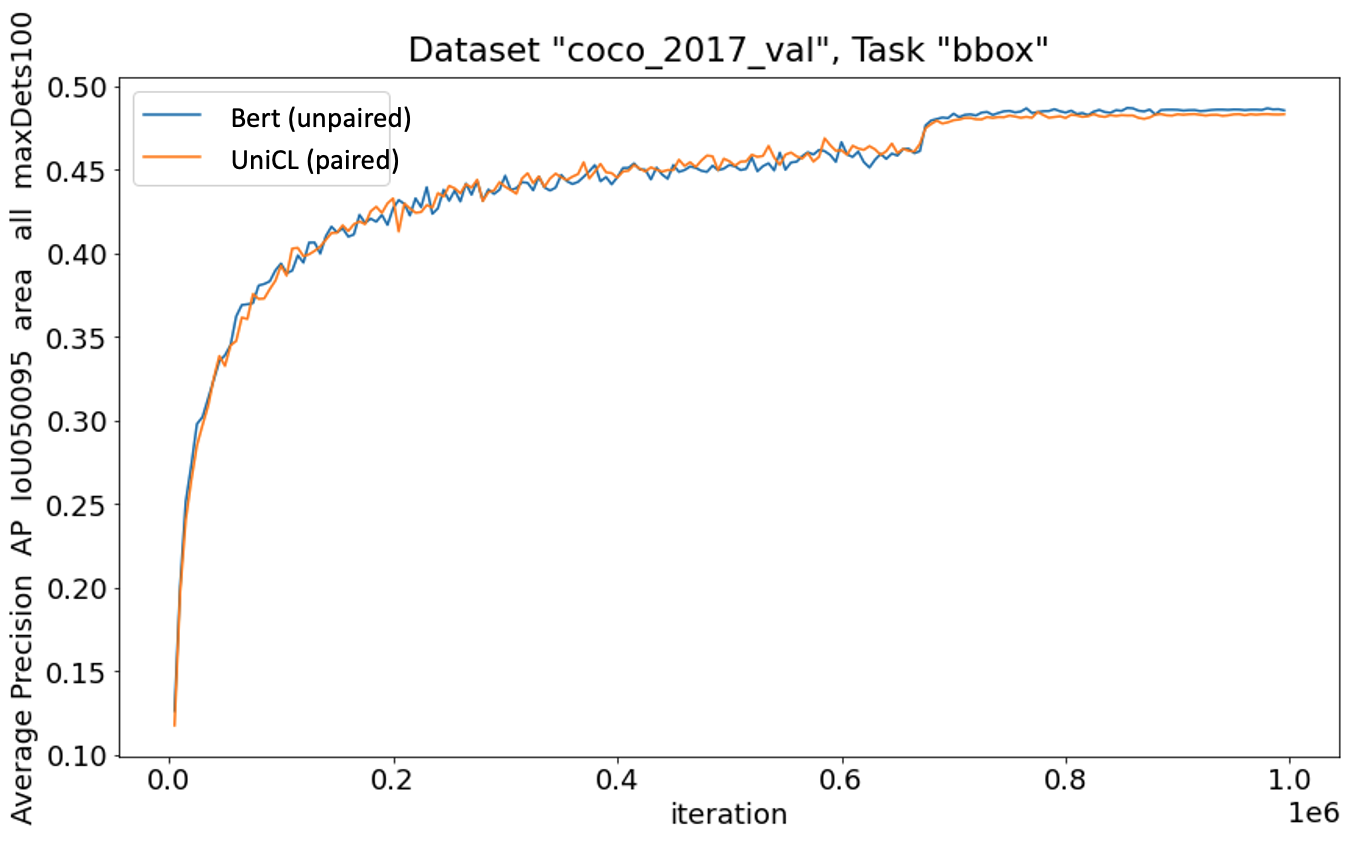} 
\caption{GLIP-B with image encoder initialized from UniCL pre-trained image encoder, but with different language encoder initialization. Blue: language encoder initialized by Bert-Base, thus un-paired image-language pre-trained encoders. Yellow: language encoder initialized from UniCL pre-trained language encoder, thus paired UniCL pre-trained image-language encoders. From the results, we can see that the COCO zero-shot transfer results from two initializations are nearly the same. Similar results have been observed for other metrics, i.e., LVIS zero-shot AP, ODinW benchmark, and Flickr30k grounding performance.}
\label{fig:ablation_language_encoder}
\end{figure}

For \ourT, we use the ImageNet pre-trained Swin-Transformer to initialize the image encoder and BERT-base-uncased to initialize the language encoder. For \ourB, we use the pre-trained paired image-language encoder from UniCL (CLIP-like pre-training, \url{https://github.com/microsoft/UniCL}) for initialization. We did an ablation study on the different language encoders (UniCL vs. BERT) and found that their results are nearly the same, as shown in Figure ~\ref{fig:ablation_language_encoder}. 
Therefore, UniCL initialization does not skew the good localization performance. The main reason for us to keep the UniCL(CLIP-like) language encoder is due to its Pre-LayerNorm \cite{xiong2020layer} operation. We find the UniCL(CLIP-like) language encoder with Pre-LayerNorm is more stable during the training compared with BERT, which uses Post-LayerNorm.

\section{More analysis on pre-training data} 
\label{sec:pre_data}

\begin{figure}[tb!]
\centering
\includegraphics[width=\textwidth]{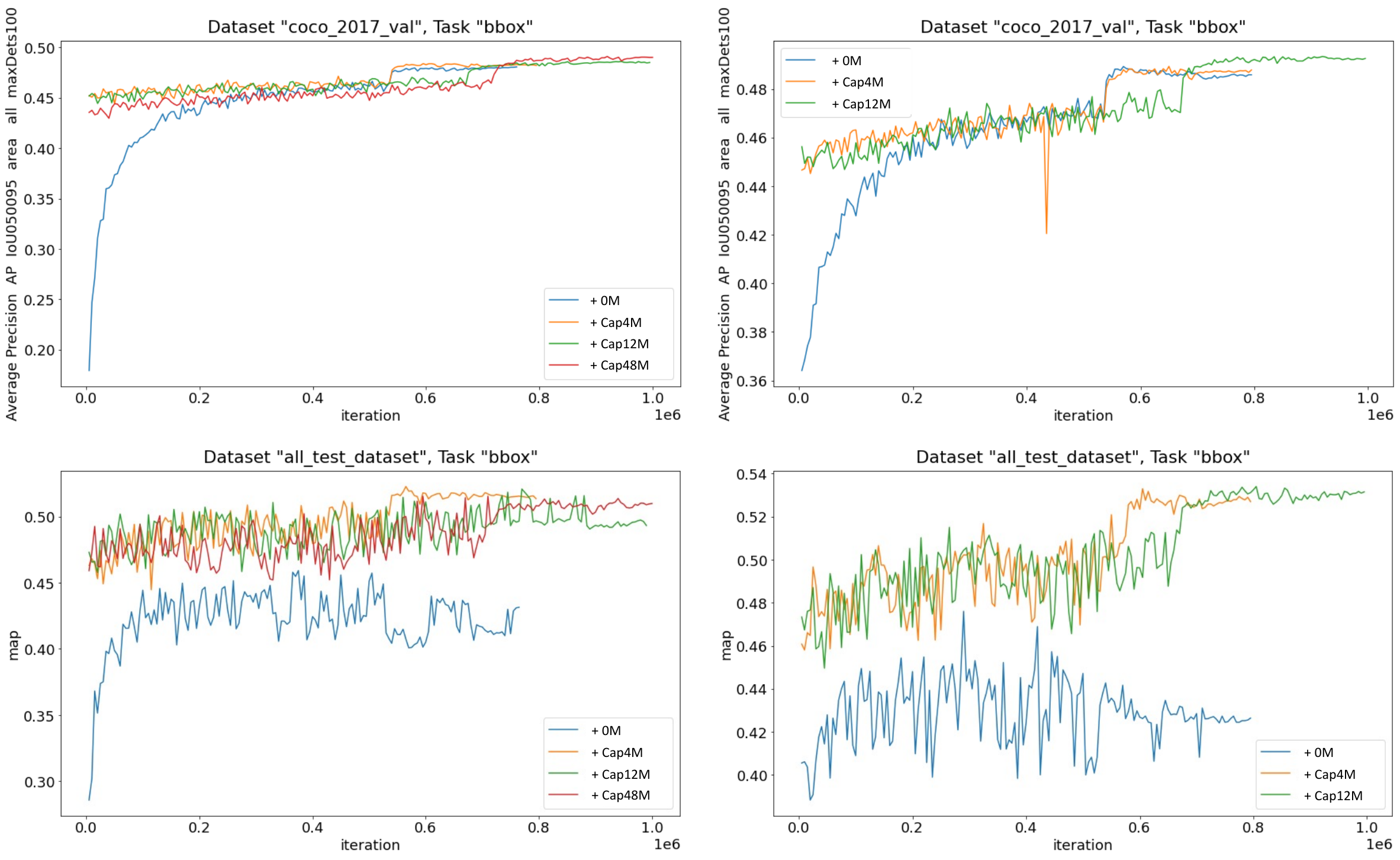} \\
\includegraphics[width=\textwidth]{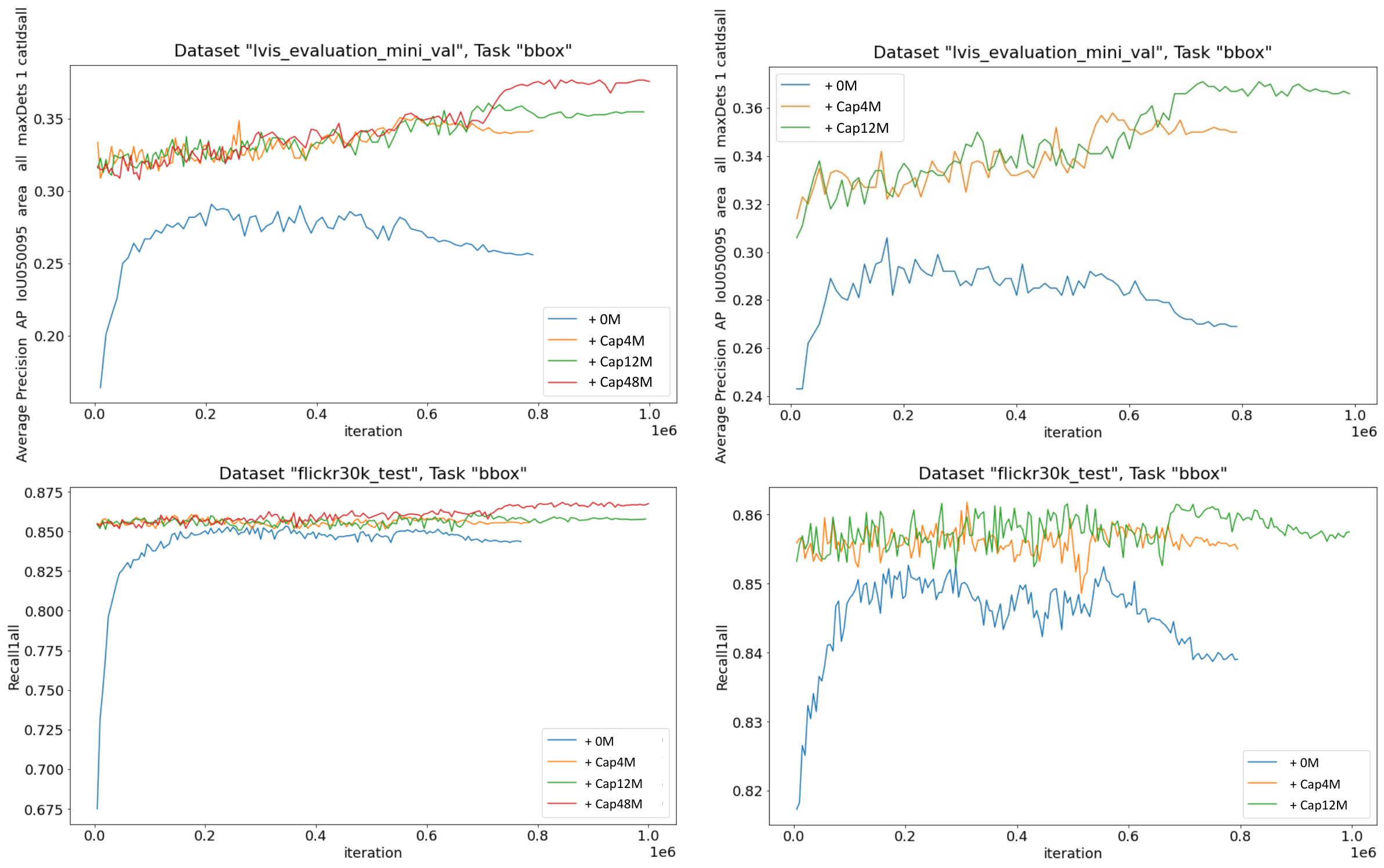} \\
\caption{Pre-train data scale up on Base-scale model. Left: GLIP, Right: GLIPv2; Row 1: COCO minival, Row 2: ODinW test split, Row 3: LVIS minival, Row 4: Flick30K test.}
\label{fig:data_allcheckpoints}
\end{figure}

Table ~\ref{tab:data_scaleup} in the main paper reports the last checkpoint results on \our when we do the scaling up of pre-training data. As more weak image-text pair data (Cap) is involved in our training, it benefits both standard/in-domain (i.e., COCO, Flickr30K) and large-domain gap (i.e., ODinW, LVIS) tasks. Further adding the inter-image region-word contrastive helps when we are fixing the data at the same scale. For large-domain gap tasks, adding the inter-image region-word contrastive loss will further boost the model to learn better representation. To learn more scaling-up effects on various tasks under all the checkpoints for GLIP and \our, see Figure ~\ref{fig:data_allcheckpoints}. Given the considerable amount of improvement of \our when the number of caption data increases from 0M to 12M, we hypothesize that it has potential to further grow by training on even larger-scale web image-text pairs. 

\section{Experiments on grounded image captioning}
\label{sec:grounded_caption}

The grounded captioning task requires the model to generate an image caption and also ground predicted phrases to object regions. The final predictions consist of (1) the text captions (2) predicted object regions, and (3) the grounding correspondence between the phrases and regions. Following the established benchmarks \cite{ma2020learning, zhou2020unified}, we evaluate the caption metrics on COCO Captions and report the grounding metrics on Flick30K, as shown in Table ~\ref{tab:grounded_caption}. 

\begin{table}[ht!]
    \centering
    \resizebox{0.8\linewidth}{!}{
    \begin{tabular}{l|ccc|ccc}
    \toprule
      \multirow{2}{*}{Model} & \multicolumn{3}{c|}{COCO Caption} & \multicolumn{3}{c}{Flickr30K Grounding} \\
         & B@4 & CIDEr & SPICE & R@1 & R@5 & R@10 \\
        \midrule
        No Pretrain & 35.4 & 115.3 & 21.2 & 77.0 & 92.9 & 95.7 \\
        + $L_{\text{mlm}}$ & 33.4 & 107.6 & 19.9 & 70.9 & 90.0 & 93.2 \\
        + $L_{\text{loc}} + L_{\text{intra}} + L_{\text{inter}}$ & 36.6 & 120.3 & 21.6 & 80.8 & 94.9 & 96.7 \\
        \ourT & 36.5 & 119.8 & 21.6 & 80.8 & 94.4 & 96.5 \\
        \ourB & 37.4 & 123.0 & 21.9 & 81.0 & 94.5 & 96.5 \\
        \bottomrule
    \end{tabular}
    }
    \caption{Grounded image captioning results on the COCO Caption, and Flickr30K Entities. We report BLEU@4, CIDer, and SPICE metrics for caption evaluation, and we use R@1, R@5, R@10 for grounding evaluation. }
    \label{tab:grounded_caption}
\end{table}

\section{Inference speed}
\label{sec:inference_speed}

We test the inference speed for \our on V100 with batch size 1 and show its comparison to MDETR, as shown in Table ~\ref{table:inference_speed}.





\begin{table}[ht!]
\caption{Model inference speed on various tasks. We report FPS, which is the number of images processed per second per GPU (higher is better).}
\label{table:inference_speed}
\begin{center}
\resizebox{\linewidth}{!}{

\begin{tabular}{l|c|c|c}
\toprule
Model 
 & Object Detection (COCO) & Phrase Grounding (Flick30K) & Referring Expression Segmentation (PhraseCut) \\
\midrule
MDETR R101 ~\cite{kamath2021mdetr} & -- & 9.31 & 3.80 \\
MDETR EffB3 ~\cite{kamath2021mdetr} & -- & 11.20 & 3.98 \\
MDETR EffB5 ~\cite{kamath2021mdetr} & -- & 9.15 & -- \\
\midrule \midrule
GLIPv2-T & 4.12 & 3.74 & 2.26 \\
GLIPv2-B & 3.01 & 3.23 & 2.39 \\
GLIPv2-H & 1.21 & 1.13 & 0.89 \\
\bottomrule
\end{tabular}

}
\end{center}
\end{table}

\section{Figure Reference}
\label{sec:figure_source}

We provided the original sources of the images that are used in our paper in the following. All datasets above were collected by the creators (cited) and consent for any personally identifiable information (PII) was ascertained by the authors where necessary.

Figure ~\ref{fig:unfied_model} in the main paper - The top left and the bottom middle figures are the 281759.jpg in COCO val set; The left right images are (from top to down: (1) 2588.jpg in ODinW Aquarium test set. (2) 13923.jpg in LVIS val set. (3) 132690.jpg in VQA2.0 val set (question id is 132690002). (4) 462565.jpg in COCO Caption val set. 

Figure ~\ref{fig:intra_inter_loss} in the main paper - The top left figure is the 209297.jpg in COCO train set; The bottom left figure is the 9378.jpg in COCO val set. 

Figure ~\ref{fig:big_unfied_model} in the Appendix - Same as Figure ~\ref{fig:unfied_model}. The top left and the bottom middle figures are the 281759.jpg in COCO val set.

Figure ~\ref{fig:examples_1} in the Appendix - Row 1 (from left to right): (1) 439715.jpg in COCO val set. (2) 6471.jpg in COCO val set. (3) 13923.jpg in COCO val set; Row 2: (1) 5521996.jpg in Flickr30K val set. (2) 764507.jpg in Flickr30K val set. (3) 7520721.jpg in Flick30K val set; Row 3: (1) 2588.jpg in ODinW Aquarium test set. (2) 143.jpg in Thermal val set. (3) ck0l9j6n6oqjo0848ps5blk3b.jpg in WildFire val set. 

Figure ~\ref{fig:examples_2} in the Appendix - Row 1 (from left to right): (1) 13923.jpg in COCO val set. (2) 6471.jpg in COCO val set. (3) 7574.jpg in COCO val set; Row 2: (1) 117320.jpg in LVIS val set. (2) 2587.jpg in LVIS val set. (3) 211120.jpg in LVIS val set; Row 3: (1) 4744.jpg in PhraseCut test set. (2) 4744.jpg in PhraseCut val set. (3) 567.jpg in PhraseCut train set.

Figure ~\ref{fig:examples_3} in the Appendix - Row 1 (from left to right): (1) 486.jpg in VQA2.0 val set (question id is 486002). (2) 262746.jpg in VQA2.0 val set (question id is 262746002). (3) 132690.jpg in VQA2.0 val set (question id is 132690002); Row 2: (1) 391895.jpg in COCO Caption val set. (2) 462565.jpg in COCO Caption val set. (3) 579056.jpg in COCO Caption val set.

\section{All results for ODinW}
\label{sec:all_results}

We report the per-dataset performance under 0,1,3,5,10-shot and full data as well as prompt tuning, and full-model tuning in Table ~\ref{table:zero_shot_full} and Table ~\ref{table:perdataset_all_1} (on the next page).

\begin{table}[ht]
\caption{Zero-shot performance on 13 ODinW datasets.}
\label{table:zero_shot_full}
\begin{center}
\resizebox{\linewidth}{!}{
\begin{tabular}{l@{\hskip9pt}| 
l@{\hskip9pt}l@{\hskip9pt}l@{\hskip9pt} 
l@{\hskip9pt}l@{\hskip9pt}l@{\hskip9pt}
l@{\hskip9pt}l@{\hskip9pt}l@{\hskip9pt}l@{\hskip9pt}
l@{\hskip9pt}l@{\hskip9pt}l@{\hskip9pt}l@{\hskip9pt}l@{\hskip9pt}l@{\hskip9pt}l}
\toprule

Model  & \small{PascalVOC} &
\small{AerialDrone} & 
\small{Aquarium} &
\small{Rabbits} &
\small{EgoHands} &
\small{Mushrooms} &
\small{Packages} &
\small{Raccoon} &
\small{Shellfish} &
\small{Vehicles} &
\small{Pistols} &
\small{Pothole} &
\small{Thermal} & 
Avg
\\
\midrule
 GLIP-T 
 & 56.2
& 12.5
& 18.4
& 70.2
& 50.0
& 73.8
& 72.3
& 57.8
& 26.3
& 56.0
& 49.6
& 17.7
& 44.1
  & 46.5
  
\\

 GLIP-L  
 & 61.7
& 7.1
& 26.9
& 75.0
& 45.5
& 49.0
& 62.8
& 63.3
& 68.9
& 57.3
& 68.6
& 25.7
& 66.0
& 52.1
\\
\midrule \midrule

 GLIPv2-T  
 & 57.6
& 10.5
& 18.4
& 71.4
& 52.7
& 77.7
& 67.7
& 58.8
& 27.8
& 55.6
& 60.1
& 20.0
& 52.4
& 48.5
\\

 GLIPv2-B  
 & 62.8
& 8.6
& 18.9
& 73.7
& 50.3
& 83.0
& 68.6
& 61.6
& 56.0
& 53.8
& 67.8
& 32.6
& 53.8
& 54.2
\\

 GLIPv2-H  
 & 66.3
& 10.9
& 30.4
& 74.6
& 55.1
& 52.1
& 71.3
& 63.8
& 66.2
& 57.2
& 66.4
& 33.8
& 73.3
& 55.5
\\

\bottomrule
\end{tabular}
}
\end{center}
\end{table}

\begin{table}[ht]
\caption{Per-dataset performance of DyHead, GLIP-T,  GLIP-L, and GLIPv2-T, GLIPv2-B and GLIPv2-H. For PascalVOC, we report the mAP (IoU=0.50:0.95) using the COCO evaluation script, to be consistent with other 12 datasets. ``Prompt'' denotes prompt tuning. ``Full'' denotes full-model tuning.}
\label{table:perdataset_all_1}
\begin{center}
\resizebox{\linewidth}{!}{
\begin{tabular}{l@{\hskip9pt} 
c@{\hskip9pt}c@{\hskip9pt}|l@{\hskip9pt} 
l@{\hskip9pt}l@{\hskip9pt}l@{\hskip9pt}
l@{\hskip9pt}l@{\hskip9pt}l@{\hskip9pt}l@{\hskip9pt}
l@{\hskip9pt}l@{\hskip9pt}l@{\hskip9pt}l@{\hskip9pt}l@{\hskip9pt}l}
\toprule

Model & Shot & Tune & \small{PascalVOC} &
\small{AerialDrone} & 
\small{Aquarium} &
\small{Rabbits} &
\small{EgoHands} &
\small{Mushrooms} &
\small{Packages} &
\small{Raccoon} &
\small{Shellfish} &
\small{Vehicles} &
\small{Pistols} &
\small{Pothole} &
\small{Thermal} & 
Avg
\\\midrule

 DyHead \scriptsize{O365} & 1 & Full
& 25.8\std{3.0}
& 16.5\std{1.8}
& 15.9\std{2.7}
& 55.7\std{6.0}
& 44.0\std{3.6}
& 66.9\std{3.9}
& 54.2\std{5.7}
& 50.7\std{7.7}
& 14.1\std{3.6}
& 33.0\std{11.0}
& 11.0\std{6.5}
& 8.2\std{4.1}
& 43.2\std{10.0}
  & 33.8\std{3.5}
\\
 DyHead \scriptsize{O365} & 3 & Full
& 40.4\std{1.0}
& 20.5\std{4.0}
& 26.5\std{1.3}
& 57.9\std{2.0}
& 53.9\std{2.5}
& 76.5\std{2.3}
& 62.6\std{13.3}
& 52.5\std{5.0}
& 22.4\std{1.7}
& 47.4\std{2.0}
& 30.1\std{6.9}
& 19.7\std{1.5}
& 57.0\std{2.3}
  & 43.6\std{1.0}
\\
 DyHead \scriptsize{O365} & 5 & Full
& 43.5\std{1.0}
& 25.3\std{1.8}
& 35.8\std{0.5}
& 63.0\std{1.0}
& 56.2\std{3.9}
& 76.8\std{5.9}
& 62.5\std{8.7}
& 46.6\std{3.1}
& 28.8\std{2.2}
& 51.2\std{2.2}
& 38.7\std{4.1}
& 21.0\std{1.4}
& 53.4\std{5.2}
  & 46.4\std{1.1}
\\
 DyHead \scriptsize{O365} & 10 & Full
& 46.6\std{0.3}
& 29.0\std{2.8}
& 41.7\std{1.0}
& 65.2\std{2.5}
& 62.5\std{0.8}
& 85.4\std{2.2}
& 67.9\std{4.5}
& 47.9\std{2.2}
& 28.6\std{5.0}
& 53.8\std{1.0}
& 39.2\std{4.9}
& 27.9\std{2.3}
& 64.1\std{2.6}
  & 50.8\std{1.3}
\\
 DyHead \scriptsize{O365} & All & Full
& 53.3 
& 28.4 
& 49.5 
& 73.5 
& 77.9 
& 84.0 
& 69.2 
& 56.2 
& 43.6 
& 59.2 
& 68.9 
& 53.7 
& 73.7 
  & 60.8
\\
\midrule
\midrule
 GLIP-T & 1 & Prompt
& 54.4\std{0.9}
& 15.2\std{1.4}
& 32.5\std{1.0}
& 68.0\std{3.2}
& 60.0\std{0.7}
& 75.8\std{1.2}
& 72.3\std{0.0} 
& 54.5\std{3.9}
& 24.1\std{3.0}
& 59.2\std{0.9}
& 57.4\std{0.6}
& 18.9\std{1.8}
& 56.9\std{2.7}
  & 49.9\std{0.6}
\\ 
 
 GLIP-T & 3 & Prompt
& 56.8\std{0.8}
& 18.9\std{3.6}
& 37.6\std{1.6}
& 72.4\std{0.5}
& 62.8\std{1.3}
& 85.4\std{2.8}
& 64.5\std{4.6}
& 69.1\std{1.8}
& 22.0\std{0.9}
& 62.7\std{1.1}
& 56.1\std{0.6}
& 25.9\std{0.7}
& 63.8\std{4.8}
  & 53.7\std{1.3}
\\ 
 
 GLIP-T & 5 & Prompt
& 58.5\std{0.5}
& 18.2\std{0.1}
& 41.0\std{1.2}
& 71.8\std{2.4}
& 65.7\std{0.7}
& 87.5\std{2.2}
& 72.3\std{0.0} 
& 60.6\std{2.2}
& 31.4\std{4.2}
& 61.0\std{1.8}
& 54.4\std{0.6}
& 32.6\std{1.4}
& 66.3\std{2.8}
  & 55.5\std{0.5}
\\ 
 
 GLIP-T & 10 & Prompt
& 59.7\std{0.7}
& 19.8\std{1.6}
& 44.8\std{0.9}
& 72.1\std{2.0}
& 65.9\std{0.6}
& 87.4\std{1.1}
& 72.3\std{0.0} 
& 57.5\std{1.2}
& 30.0\std{1.4}
& 62.1\std{1.4}
& 57.8\std{0.9}
& 33.5\std{0.1}
& 73.1\std{1.4}
  & 56.6\std{0.2}
\\ 
 
 GLIP-T & All & Prompt
& 66.4 
& 27.6 
& 50.9 
& 70.6 
& 73.3 
& 88.1 
& 67.7 
& 64.0 
& 40.3 
& 65.4 
& 68.3 
& 50.7 
& 78.5 
  & 62.4
\\

\midrule
 GLIP-T & 1 & Full
 & 54.8\std{2.0}
& 18.4\std{1.0}
& 33.8\std{1.1}
& 70.1\std{2.9}
& 64.2\std{1.8}
& 83.7\std{3.0}
& 70.8\std{2.1}
& 56.2\std{1.8}
& 22.9\std{0.2}
& 56.6\std{0.5}
& 59.9\std{0.4}
& 18.9\std{1.3}
& 54.5\std{2.7}
  & 51.1\std{0.1}
 
 \\

 GLIP-T & 3 & Full
 & 58.1\std{0.5}
& 22.9\std{1.3}
& 40.8\std{0.9}
& 65.7\std{1.6}
& 66.0\std{0.2}
& 84.7\std{0.5}
& 65.7\std{2.8}
& 62.6\std{1.4}
& 27.2\std{2.7}
& 61.9\std{1.8}
& 60.7\std{0.2}
& 27.1\std{1.2}
& 70.4\std{2.5}
  & 54.9\std{0.2}
 
 \\
 GLIP-T & 5 & Full
 & 59.5\std{0.4}
& 23.8\std{0.9}
& 43.6\std{1.4}
& 68.7\std{1.3}
& 66.1\std{0.6}
& 85.4\std{0.4}
& 72.3\std{0.0} 
& 62.1\std{2.0}
& 27.3\std{1.2}
& 61.0\std{1.8}
& 62.7\std{1.6}
& 34.5\std{0.5}
& 66.6\std{2.3}
  & 56.4\std{0.4}
 
 \\
 GLIP-T & 10 & Full
 & 59.1\std{1.3}
& 26.3\std{1.1}
& 46.3\std{1.6}
& 67.3\std{1.5}
& 67.1\std{0.7}
& 87.8\std{0.5}
& 72.3\std{0.0} 
& 57.7\std{1.7}
& 34.6\std{1.7}
& 65.4\std{1.4}
& 61.6\std{1.0}
& 39.3\std{1.0}
& 74.7\std{2.3}
  & 58.4\std{0.2}
 
 \\
 GLIP-T & All & Full
 & 62.3 
& 31.2 
& 52.5 
& 70.8 
& 78.7 
& 88.1 
& 75.6 
& 61.4 
& 51.4 
& 65.3 
& 71.2 
& 58.7 
& 76.7 
  & 64.9
 \\
 \midrule
 \midrule
 
 GLIP-L & 1 & Prompt
 & 62.8\std{0.4}
& 18.0\std{1.8}
& 37.4\std{0.3}
& 71.9\std{2.4}
& 68.9\std{0.1}
& 81.8\std{3.4}
& 65.0\std{2.8}
& 63.9\std{0.4}
& 70.2\std{1.2}
& 67.0\std{0.4}
& 69.3\std{0.1}
& 27.6\std{0.4}
& 69.8\std{0.6}
  & 59.5\std{0.4}
 
 \\ 
 GLIP-L & 3 & Prompt
 & 65.0\std{0.5}
& 21.4\std{1.0}
& 43.6\std{1.1}
& 72.9\std{0.7}
& 70.4\std{0.1}
& 91.4\std{0.7}
& 57.7\std{3.7}
& 70.7\std{1.1}
& 69.7\std{0.9}
& 62.6\std{0.8}
& 67.7\std{0.4}
& 36.2\std{1.1}
& 68.8\std{1.5}
  & 61.4\std{0.3}
 
 \\ 
 GLIP-L & 5 & Prompt
 
 & 65.6\std{0.3}
& 19.9\std{1.6}
& 47.7\std{0.7}
& 73.7\std{0.7}
& 70.6\std{0.3}
& 86.8\std{0.5}
& 64.6\std{0.7}
& 69.4\std{3.3}
& 68.0\std{1.3}
& 67.8\std{1.5}
& 68.3\std{0.3}
& 36.6\std{1.6}
& 71.9\std{0.6}
  & 62.4\std{0.5}
 \\ 
 GLIP-L & 10 & Prompt
 & 65.9\std{0.2}
& 23.4\std{2.6}
& 50.3\std{0.7}
& 73.6\std{0.7}
& 71.8\std{0.3}
& 86.5\std{0.3}
& 70.5\std{1.1}
& 69.0\std{0.5}
& 69.4\std{2.4}
& 70.8\std{1.2}
& 68.8\std{0.6}
& 39.3\std{0.9}
& 74.9\std{2.1}
  & 64.2\std{0.4}
 
 \\ 
 GLIP-L & All & Prompt
 
 & 72.9 
& 23.0 
& 51.8 
& 72.0 
& 75.8 
& 88.1 
& 75.2 
& 69.5 
& 73.6 
& 72.1 
& 73.7 
& 53.5 
& 81.4 
  & 67.9\std{0.0}

 \\ 
\midrule
 GLIP-L & 1 & Full
 & 64.8\std{0.6}
& 18.7\std{0.6}
& 39.5\std{1.2}
& 70.0\std{1.5}
& 70.5\std{0.2}
& 69.8\std{18.0}
& 70.6\std{4.0}
& 68.4\std{1.2}
& 71.0\std{1.3}
& 65.4\std{1.1}
& 68.1\std{0.2}
& 28.9\std{2.9}
& 72.9\std{4.7}
  & 59.9\std{1.4}
 
 \\ 
 GLIP-L & 3 & Full
 
 & 65.6\std{0.6}
& 22.3\std{1.1}
& 45.2\std{0.4}
& 72.3\std{1.4}
& 70.4\std{0.4}
& 81.6\std{13.3}
& 71.8\std{0.3}
& 65.3\std{1.6}
& 67.6\std{1.0}
& 66.7\std{0.9}
& 68.1\std{0.3}
& 37.0\std{1.9}
& 73.1\std{3.3}
  & 62.1\std{0.7}
 \\ 
 GLIP-L & 5 & Full
 
 & 66.6\std{0.4}
& 26.4\std{2.5}
& 49.5\std{1.1}
& 70.7\std{0.2}
& 71.9\std{0.2}
& 88.1\std{0.0} 
& 71.1\std{0.6}
& 68.8\std{1.2}
& 68.5\std{1.7}
& 70.0\std{0.9}
& 68.3\std{0.5}
& 39.9\std{1.4}
& 75.2\std{2.7}
  & 64.2\std{0.3}

 \\ 
 GLIP-L & 10 & Full
 & 66.4\std{0.7}
& 32.0\std{1.4}
& 52.3\std{1.1}
& 70.6\std{0.7}
& 72.4\std{0.3}
& 88.1\std{0.0}
& 67.1\std{3.6}
& 64.7\std{3.1}
& 69.4\std{1.4}
& 71.5\std{0.8}
& 68.4\std{0.7}
& 44.3\std{0.6}
& 76.3\std{1.1}
  & 64.9\std{0.7}
 
 \\
 GLIP-L & All & Full
 & 69.6 
& 32.6 
& 56.6 
& 76.4 
& 79.4 
& 88.1 
& 67.1 
& 69.4 
& 65.8 
& 71.6 
& 75.7 
& 60.3 
& 83.1 
  & 68.9
 \\
 
\midrule
\midrule

GLIPv2-T & 1 & Prompt
 & 51.2\std{0.3}
& 17.7\std{1.2}
& 34.2\std{0.1}
& 68.7\std{1.2}
& 67.3\std{0.9}
& 83.7\std{2.1}
& 68.1\std{1.7}
& 53.4\std{0.2}
& 30.0\std{0.9}
& 59.0\std{0.1}
& 60.0\std{0.3}
& 21.9\std{0.2}
& 66.5\std{0.7}
  & 52.4\std{0.5}
 
 \\ 
 GLIPv2-T & 3 & Prompt
 & 66.6\std{0.2}
& 11.5\std{0.7}
& 37.2\std{1.0}
& 71.7\std{0.3}
& 70.1\std{0.4}
& 45.7\std{0.1}
& 57.7\std{1.2}
& 69.7\std{1.5}
& 42.7\std{0.4}
& 67.5\std{0.9}
& 65.6\std{1.0}
& 36.7\std{1.2}
& 69.2\std{1.2}
  & 55.6\std{0.4}
 
 \\ 
 GLIPv2-T & 5 & Prompt
 
 & 58.9\std{1.2}
& 17.4\std{0.6}
& 42.8\std{0.4}
& 72.6\std{0.5}
& 66.1\std{0.2}
& 84.9\std{0.8}
& 69.7\std{0.6}
& 65.5\std{2.1}
& 35.6\std{0.8}
& 62.8\std{0.9}
& 59.8\std{0.2}
& 35.5\std{0.9}
& 74.4\std{0.2}
  & 57.4\std{0.4}
 \\ 
 GLIPv2-T & 10 & Prompt
 & 59.9\std{0.4}
& 21.6\std{2.0}
& 43.7\std{0.3}
& 74.3\std{0.4}
& 68.2\std{0.7}
& 88.1\std{0.1}
& 72.0\std{0.9}
& 60.0\std{0.4}
& 35.6\std{1.2}
& 66.1\std{0.6}
& 61.0\std{0.3}
& 42.8\std{0.4}
& 70.9\std{3.2}
  & 58.8\std{0.5}
 
 \\ 
 GLIPv2-T & All & Prompt
 
 & 67.4 
& 22.3
& 50.5 
& 74.3 
& 73.4 
& 85.5 
& 74.7 
& 65.8 
& 53.7 
& 67.4 
& 68.9 
& 52.3 
& 83.7 
  & 64.8\std{0.0}

 \\ 
\midrule
 GLIPv2-T & 1 & Full
 & 64.8\std{0.6}
& 18.7\std{0.6}
& 39.5\std{1.2}
& 70.0\std{1.5}
& 70.5\std{0.2}
& 69.8\std{18.0}
& 70.6\std{4.0}
& 68.4\std{1.2}
& 71.0\std{1.3}
& 65.4\std{1.1}
& 68.1\std{0.2}
& 28.9\std{2.9}
& 72.9\std{4.7}
  & 52.8\std{1.4}
 
 \\ 
 GLIPv2-T & 3 & Full
 
 & 53.9\std{0.1}
& 17.8\std{0.7}
& 42.7\std{1.1}
& 73.1\std{1.0}
& 65.9\std{0.2}
& 84.7\std{3.4}
& 69.7\std{0.8}
& 60.7\std{1.3}
& 28.8\std{0.8}
& 61.7\std{1.3}
& 60.6\std{0.2}
& 35.5\std{0.4}
& 68.3\std{1.7}
  & 55.6\std{0.7}
 \\ 
 GLIPv2-T & 5 & Full
 
 & 58.9\std{0.2}
& 17.4\std{1.1}
& 42.8\std{1.3}
& 72.6\std{0.7}
& 66.1\std{0.6}
& 84.9\std{0.9} 
& 69.7\std{0.3}
& 65.5\std{1.0}
& 35.6\std{0.9}
& 62.8\std{0.3}
& 59.8\std{0.2}
& 35.5\std{1.2}
& 74.4\std{2.1}
  & 57.4\std{0.4}

 \\ 
 GLIPv2-T & 10 & Full
 & 57.6\std{1.0}
& 27.6\std{1.2}
& 49.1\std{1.0}
& 70.4\std{0.5}
& 69.2\std{0.2}
& 88.1\std{0.0}
& 73.1\std{2.3}
& 58.0\std{2.8}
& 42.9\std{1.2}
& 64.8\std{0.2}
& 62.1\std{0.9}
& 39.9\std{0.4}
& 71.6\std{0.8}
  & 59.7\std{0.3}
 
 \\
 GLIPv2-T & All & Full
 & 66.4
& 30.2 
& 52.5 
& 74.8 
& 80.0 
& 88.1 
& 74.3 
& 63.7 
& 54.4 
& 63.0 
& 73.0 
& 60.1 
& 83.5
  & 66.5
 \\
\midrule
\midrule

GLIPv2-B & 1 & Prompt
 & 68.7\std{0.1}
& 19.9\std{0.3}
& 38.4\std{0.8}
& 68.5\std{1.0}
& 68.6\std{0.8}
& 87.7\std{3.0}
& 69.3\std{1.7}
& 68.5\std{0.4}
& 55.2\std{0.3}
& 65.7\std{0.7}
& 67.2\std{0.1}
& 34.8\std{0.8}
& 69.6\std{0.4}
  & 60.4\std{0.3}
 
 \\ 
 GLIPv2-B & 3 & Prompt
 & 67.2\std{0.6}
& 22.2\std{0.3}
& 46.5\std{0.9}
& 71.2\std{0.8}
& 70.9\std{0.1}
& 86.9\std{0.2}
& 67.7\std{1.8}
& 63.7\std{2.3}
& 46.9\std{0.8}
& 68.1\std{0.4}
& 67.4\std{0.9}
& 47.9\std{1.0}
& 78.9\std{1.7}
  & 62.0\std{0.5}
 
 \\ 
 GLIPv2-B & 5 & Prompt
 
 & 68.9\std{1.0}
& 25.7\std{0.4}
& 50.5\std{0.9}
& 73.8\std{1.5}
& 69.7\std{0.6}
& 84.9\std{0.3}
& 69.3\std{0.7}
& 65.8\std{1.6}
& 65.7\std{1.0}
& 69.2\std{0.3}
& 67.5\std{0.7}
& 34.0\std{0.2}
& 73.1\std{0.6}
  & 62.9\std{0.4}
 \\ 
 GLIPv2-B & 10 & Prompt
 & 69.4\std{0.7}
& 21.8\std{1.3}
& 48.7\std{0.2}
& 71.3\std{0.2}
& 71.0\std{0.7}
& 88.1\std{0.4}
& 68.6\std{0.7}
& 73.5\std{0.3}
& 61.5\std{1.9}
& 69.3\std{0.2}
& 68.6\std{0.7}
& 41.3\std{0.2}
& 75.2\std{1.3}
  & 63.8\std{0.3}
 
 \\ 
 GLIPv2-B & All & Prompt
 
 & 71.9 
& 26.1
& 50.6 
& 74.5 
& 73.5 
& 86.9 
& 74.9 
& 71.0 
& 71.6
& 71.0 
& 72.4 
& 50.2 
& 80.5 
  & 67.3\std{0.0}

 \\ 
\midrule
 GLIPv2-B & 1 & Full
 & 67.8\std{0.4}
& 18.7\std{0.3}
& 44.2\std{0.9}
& 71.4\std{0.3}
& 70.4\std{1.2}
& 87.9\std{7.3}
& 66.1\std{2.4}
& 68.9\std{1.1}
& 60.6\std{1.6}
& 68.1\std{0.6}
& 69.0\std{0.7}
& 35.1\std{0.9}
& 68.9\std{2.1}
  & 61.2\std{0.6}
 
 \\ 
 GLIPv2-B & 3 & Full
 
 & 68.1\std{0.2}
& 25.7\std{0.4}
& 46.4\std{1.6}
& 69.8\std{1.3}
& 71.3\std{1.2}
& 88.0\std{3.4}
& 68.6\std{0.9}
& 69.8\std{1.7}
& 60.1\std{0.3}
& 68.4\std{1.9}
& 68.5\std{0.6}
& 39.8\std{0.8}
& 71.4\std{2.1}
  & 62.8\std{0.8}
 \\ 
 GLIPv2-B & 5 & Full
 
 & 68.6\std{1.0}
& 21.6\std{0.6}
& 46.7\std{0.7}
& 70.9\std{0.9}
& 71.0\std{1.2}
& 88.1\std{3.7} 
& 69.1\std{0.2}
& 71.8\std{1.0}
& 61.5\std{0.7}
& 68.7\std{0.2}
& 69.3\std{0.8}
& 40.2\std{1.0}
& 74.8\std{2.8}
  & 63.3\std{0.6}

 \\ 
 GLIPv2-B & 10 & Full
 & 67.4\std{1.3}
& 22.3\std{1.1}
& 50.5\std{0.7}
& 74.3\std{0.4}
& 73.4\std{0.4}
& 85.5\std{0.1}
& 74.7\std{0.9}
& 65.8\std{2.4}
& 53.7\std{1.1}
& 67.4\std{0.9}
& 68.9\std{0.7}
& 52.3\std{0.6}
& 83.7\std{3.2}
  & 64.6\std{0.3}
 
 \\
 GLIPv2-B & All & Full
 & 71.1
& 32.6
& 57.5 
& 73.6 
& 80.0 
& 88.1 
& 74.9 
& 68.2 
& 70.6 
& 71.2 
& 76.5 
& 58.7 
& 79.6
  & 69.4
 \\
 
 \midrule
\midrule

GLIPv2-H & 1 & Prompt
 & 68.3\std{0.6}
& 16.4\std{0.6}
& 45.8\std{0.3}
& 72.0\std{0.5}
& 67.9\std{0.9}
& 89.3\std{3.2}
& 69.3\std{1.7}
& 67.9\std{0.8}
& 66.3\std{1.9}
& 68.0\std{0.7}
& 66.8\std{0.3}
& 33.9\std{0.4}
& 70.7\std{1.5}
  & 61.4\std{0.5}
 
 \\ 
 GLIPv2-H & 3 & Prompt
 & 69.5\std{0.7}
& 25.9\std{0.2}
& 50.0\std{1.2}
& 75.4\std{1.4}
& 70.1\std{0.9}
& 85.9\std{2.5}
& 69.3\std{0.7}
& 70.8\std{1.2}
& 66.4\std{0.8}
& 68.0\std{1.2}
& 68.8\std{0.9}
& 34.0\std{0.3}
& 72.7\std{1.6}
  & 63.6\std{0.6}
 
 \\ 
 GLIPv2-H & 5 & Prompt
 
 & 69.4\std{0.7}
& 22.0\std{0.6}
& 49.1\std{0.1}
& 70.7\std{1.0}
& 73.0\std{0.5}
& 88.1\std{0.8}
& 70.3\std{0.4}
& 71.2\std{1.8}
& 62.9\std{1.4}
& 70.1\std{0.3}
& 68.3\std{0.6}
& 42.7\std{0.6}
& 74.3\std{0.5}
  & 63.9\std{0.7}
 \\ 
 GLIPv2-H & 10 & Prompt
 & 66.0\std{0.7}
& 27.5\std{1.3}
& 53.8\std{0.2}
& 74.6\std{0.2}
& 80.1\std{0.7}
& 87.4\std{0.4}
& 69.3\std{0.7}
& 66.0\std{0.3}
& 51.2\std{1.9}
& 67.2\std{0.2}
& 72.8\std{0.7}
& 58.3\std{0.2}
& 76.5\std{1.3}
  & 65.5\std{0.6}
 
 \\ 
 GLIPv2-H & All & Prompt
 
 & 71.2 
& 31.1
& 57.1 
& 75.0 
& 79.8 
& 88.1 
& 68.6 
& 68.3 
& 59.6
& 70.9
& 73.6
& 61.4
& 78.6 
  & 69.1\std{0.0}

 \\ 
\midrule
 GLIPv2-H & 1 & Full
 & 67.8\std{0.6}
& 17.3\std{0.6}
& 50.7\std{0.3}
& 63.8\std{0.5}
& 67.3\std{0.9}
& 89.4\std{3.2}
& 69.3\std{1.7}
& 68.2\std{0.8}
& 66.6\std{1.9}
& 66.8\std{0.7}
& 67.0\std{0.3}
& 34.0\std{0.4}
& 75.0\std{1.5}
  & 61.7\std{0.5}
 
 \\ 
 GLIPv2-H & 3 & Full
 
 & 62.3\std{0.2}
& 29.1\std{0.4}
& 53.8\std{1.6}
& 72.7\std{1.3}
& 78.4\std{1.2}
& 85.8\std{3.4}
& 68.6\std{0.9}
& 60.7\std{1.7}
& 43.6\std{0.3}
& 65.9\std{1.9}
& 72.2\std{0.6}
& 55.9\std{0.8}
& 81.1\std{2.1}
  & 64.1\std{0.8}
 \\ 
 GLIPv2-H & 5 & Full
 
 & 66.4\std{1.0}
& 23.4\std{0.6}
& 50.7\std{0.7}
& 73.9\std{0.9}
& 71.8\std{1.2}
& 84.2\std{3.7} 
& 71.2\std{0.2}
& 68.1\std{1.0}
& 67.4\std{0.7}
& 70.8\std{0.2}
& 65.8\std{0.8}
& 54.6\std{1.0}
& 75.6\std{2.8}
  & 64.4\std{0.6}

 \\ 
 GLIPv2-H & 10 & Full
 & 67.3\std{1.3}
& 31.6\std{1.1}
& 52.4\std{0.7}
& 71.3\std{0.4}
& 80.0\std{0.4}
& 88.1\std{0.1}
& 72.9\std{0.9}
& 56.9\std{2.4}
& 52.2\std{1.1}
& 65.4\std{0.9}
& 73.9\std{0.7}
& 61.0\std{0.6}
& 84.0\std{3.2}
  & 65.9\std{0.3}
 
 \\
 GLIPv2-H & All & Full
 & 74.4
& 36.3
& 58.7 
& 77.1 
& 79.3 
& 88.1 
& 74.3 
& 73.1 
& 70.0 
& 72.2 
& 72.5 
& 58.3 
& 81.4
  & 70.4
  
  \\

\bottomrule
\end{tabular}
}
\end{center}
\end{table}

\end{document}